\documentclass[10pt]{article}

\usepackage[utf8]{inputenc} 
\usepackage[T1]{fontenc}    
\usepackage{hyperref}       
\usepackage{url}            
\usepackage{booktabs}       
\usepackage{amsfonts}       
\usepackage{nicefrac}       
\usepackage{microtype}      
\usepackage{graphicx}
\usepackage{subfigure}
\usepackage{booktabs} 
\usepackage{xcolor}         
\usepackage{amsmath, amsthm, amsfonts}

\usepackage{algorithmic, algorithm}
\usepackage{bbm} 
\newcommand{\indicator}{\mathbbm{1}}  
\newtheorem{prop}{Proposition}

\usepackage{verbatim}
\usepackage{mathtools}

\usepackage[final,nonatbib]{neurips_2025}
\usepackage[square,numbers,sort&compress]{natbib} 

\usepackage[utf8]{inputenc} 
\usepackage[T1]{fontenc}    
\usepackage{hyperref}       
\usepackage{url}            
\usepackage{booktabs}       
\usepackage{amsfonts}       
\usepackage{nicefrac}       
\usepackage{microtype}      
\usepackage{xcolor}         

\newcommand{\vw}{\mathbf{w}}
\newcommand{\vx}{\mathbf{x}}
\newcommand{\RR}{\mathbb{R}}
\newcommand{\trp}{^\top}
\newcommand{\wstim}{w_{stim}}

\usepackage{authblk}

\author[1,*]{Yuhan Helena Liu}
\author[1]{Victor Geadah}
\author[1,*]{Jonathan Pillow}
\affil[1]{Princeton University, Princeton, NJ, USA} 
\affil[*]{Correspondence: hl7582@princeton.edu, pillow@princeton.edu} 

\title{Flexible inference for animal learning rules using neural networks}

\date{}

\begin{document}

\maketitle 

\begin{abstract}
Understanding how animals learn is a central challenge in neuroscience, with growing relevance to the development of animal- or human-aligned artificial intelligence. However, existing approaches tend to assume fixed parametric forms for the learning rule (e.g., Q-learning, policy gradient), which may not accurately describe the complex forms of learning employed by animals in realistic settings. 
Here we address this gap by developing a framework to infer learning rules directly from behavioral data collected during {\it de novo} task learning. We assume that animals follow a decision policy parameterized by a generalized linear model (GLM), and we model their learning rule---the mapping from task covariates to per-trial weight updates---using a deep neural network (DNN). This formulation allows flexible, data-driven inference of learning rules while maintaining an interpretable form of the decision policy itself. To capture more complex learning dynamics, we introduce a recurrent neural network (RNN) variant that relaxes the Markovian assumption that learning depends solely on covariates of the current trial, allowing for learning rules that integrate information over multiple trials. Simulations demonstrate that the framework can recover ground-truth learning rules. 
We applied our DNN and RNN-based methods to a large behavioral dataset from mice learning to perform a sensory decision-making task and found that they outperformed traditional RL learning rules at predicting the learning trajectories of held-out mice. The inferred learning rules exhibited reward-history–dependent learning dynamics, with larger updates following sequences of rewarded trials. Overall, these methods provide a flexible framework for inferring learning rules from behavioral data in {\it de novo} learning tasks, setting the stage for improved animal training protocols and the development of behavioral digital twins. 
\end{abstract} 

\section{Introduction} 

The study of animal learning has a long history, beginning with foundational work such as Pavlov’s classical conditioning and the Rescorla–Wagner model of associative learning~\cite{Pavlov1927,Rescorla1972}. Uncovering the mechanisms that underlie learning and decision-making is crucial for understanding and predicting animal behavior~\cite{fazzari2024animal}. This knowledge has wide-reaching applications, including benefits for human health and medicine~\cite{gnanasekar2022rodent,shaw2023human,manduca2023learning,hart2011behavioural}, informing conservation efforts by anticipating how animals respond to environmental change~\cite{wijeyakulasuriya2020machine,hou2020identification,ditria2020automating,pillai2023deep,nilsson2020simple}, and potentially guiding the development of biologically aligned AI systems, which have garnered increasing interest in recent years~\cite{sucholutsky2023getting,zador2023catalyzing}. 

While inference of learning rules is gaining traction, prior studies have key limitations. For example, many leverage reinforcement learning (RL) paradigms with normative approaches that assume reward maximization \cite{dayan2008decision,niv2009reinforcement,sutton1988time} or weight update under a parametric learning rule (e.g., Q-learning~\cite{Lak2020elife}, REINFORCE~\cite{ashwood2020inferring,geadah2025inferring}). Although these approaches have provided clear insights into animal learning, they may lack the flexibility to capture the detailed structure of real learning rules, which are often unknown and may not align with parametric forms conceived by human researchers. As a result, existing models may span only a small subset of the space of possible learning strategies animals actually use. 
Recent work has proposed more flexible quantitative models of learning based on recurrent neural networks (RNNs)~\cite{ji2023automatic,miller2023cognitive,eckstein2023predictive,dezfouli2019disentangled} or mechanisms proposed by large language models~\cite{castro2025discovering}. However, these studies have generally focused on ``bandit'' style tasks, in which animals adapt their behavior to time-varying rewards in a fixed task structure, as opposed to learning a novel task or behavior from scratch~\cite{chang2023novo,greenwell2023understanding}. \textit{De novo} learning is common in experimental neuroscience, as animals are routinely trained to acquire novel input–output mappings from scratch, yet comparatively understudied.
Finally, a variety of descriptive models have been developed to characterize trial-by-trial changes in weights that govern an animal's policy as it evolves over learning, but do not identify any rules governing these weight updates~\cite{roy2018efficient,Bruijns2023}. 

To address these shortcomings, we propose a new approach with the following \textbf{main contributions}. To our knowledge, this is the first approach to infer nonparametric, non-Markovian learning rules directly from behavior in \textit{de novo} tasks.
\begin{enumerate} 
    \item \textbf{Inferring flexible learning rules using DNNs.} We develop a deep neural network (DNN) modeling approach to infer flexible animal learning rules from \textit{de novo} task learning data (Figs.~\ref{fig:main_schematics} and~\ref{fig:main_default}). 
    \item \textbf{Non-Markovian learning rules.} Many classic learning rules are Markovian or ``memoryless'', meaning that the learning update depends only on the input, action, and reward on the current time step. Such rules, including our own DNN approach, thus fail to capture ``non-Markovian'' learning, where updates depend on multiple time steps of trial history.   
    To overcome this limitation, we incorporate a recurrent neural network (RNN) to capture arbitrary dependencies of learning on trial history (Fig.~\ref{fig:main_nonMarkovian}). 
    \item \textbf{Improved held-out prediction on real data.} We apply our methods to mouse training data from the International Brain Laboratory (IBL) and observe significantly higher log-likelihood on held-out data compared to existing approaches (Fig.~\ref{fig:main_ibl}). 
    \item \textbf{Insights into animal learning strategies.} We analyze the inferred learning rules and identify several departures from classic policy gradient learning, including non-Markovian reward history dependencies lasting multiple trials (Fig.~\ref{fig:main_ibl2}). 
\end{enumerate}

\section{Model and methods} \label{scn:methods}

\textbf{Model of decision-making}. 
We begin with a dynamic Bernoulli generalized linear model (GLM) of an animal's time-varying behavior during learning \cite{roy2018efficient, roy2021extracting, ashwood2020inferring,geadah2025inferring}. This provides an interpretable model of an animal's decision-making policy in terms of a set of weights that evolve dynamically over the course of learning. This differs from recent work in which the animal's policy is directly parametrized by an RNN \cite{ji2023automatic,miller2023cognitive}, which offers increased flexibility but may be challenging to interpret. 

\begin{figure}[t] 
    \centering
    \includegraphics[width=\textwidth]{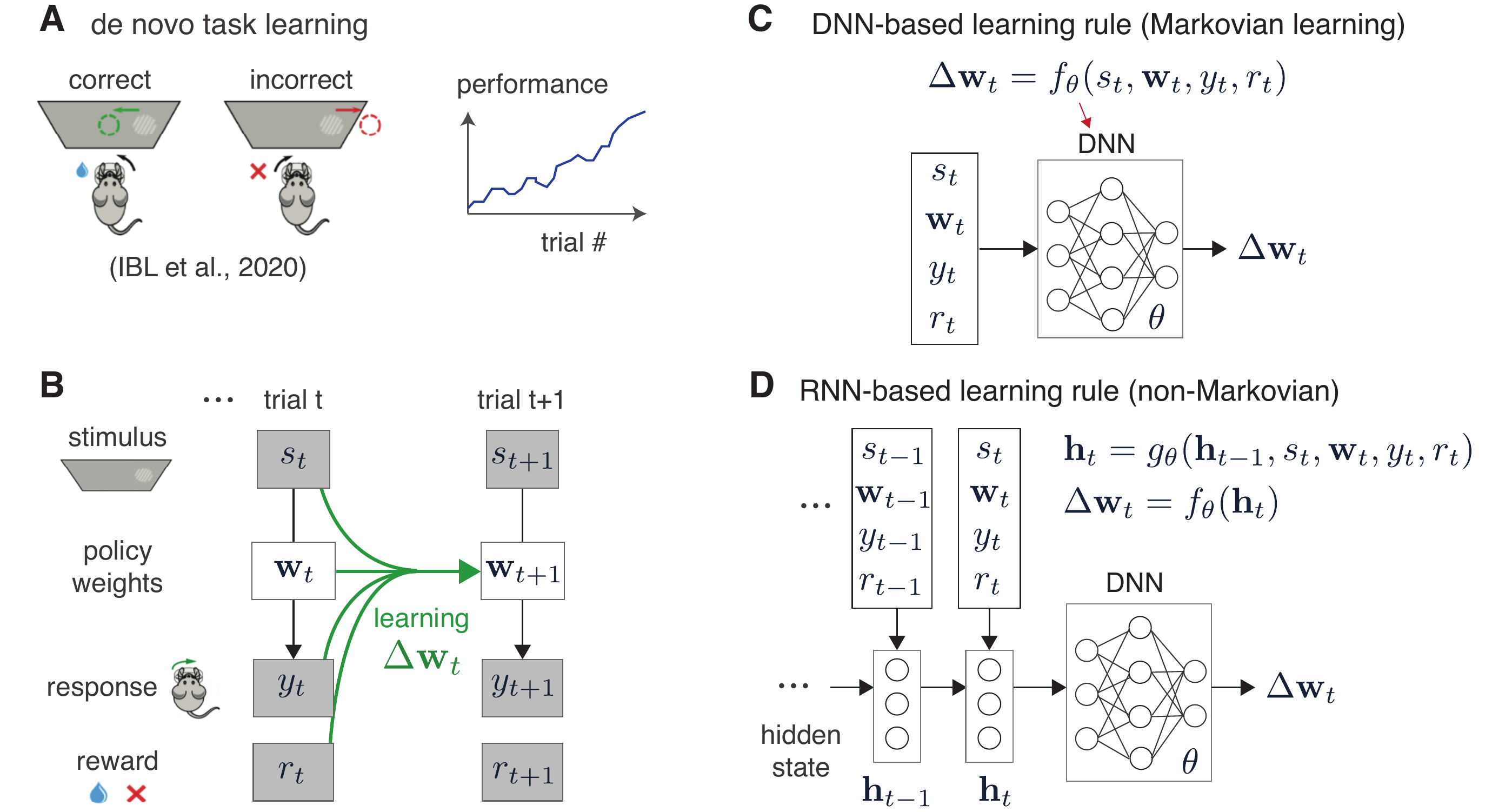}
    \caption{ {\bf Task schematic and learning rule inference methods}. \textbf{(A)} We examine learning of a sensory decision-making task, in which mice must learn to report which side of the screen contains a visual stimulus by turning a wheel \cite{international2020standardized}.  
\textbf{(B)} In our framework, decision-making is governed by the weights of a Bernoulli generalized linear model (GLM), which evolve across trials according to an unknown learning rule.
\textbf{(C-D)} To infer the learning rule, we approximate the weight update function $\Delta \vw_t$ using either (C) a deep neural network (DNN) that maps the current trial covariates to a weight change; or (D) a recurrent neural network (RNN) that integrates information from previous trials before feeding into the DNN. We optimized the neural network model parameters $\theta$ by maximizing the log-probability of the animal choice data under the dynamic Bernoulli GLM 
(see Methods and Algorithm~\ref{alg:learning-rule-inference}). We refer to our approaches as $\textbf{DNNGLM}$ or $\textbf{RNNGLM}$ depending on the model used to parametrize the learning rule.} \label{fig:main_schematics} 
\end{figure}

On trial \( t \), a sensory stimulus $s_t \in [-1,1]$ appears on the left or right side of the screen, where $s_t>0$ indicates the fractional contrast of a right-side stimulus and  $s_t<0$ indicates the (negative) fractional contrast of a left-side stimulus. The mouse makes a binary choice \( y_t \in \{0, 1\} \) by turning a wheel, where $0$ corresponds to a leftward and $1$ to a rightward choice, respectively (Fig.~\ref{fig:main_schematics}A). 
The true label of the stimulus is denoted as $z_t = \indicator[s_t > 0]$, representing the correct response according to the task.

We assume the animal’s policy on trial $t$ is governed by a weight vector \( \vw_t \in \RR^d \), which describes how task covariates, denoted \( \vx_t \in \RR^d \), influence the animal's choice. Unless otherwise specified, we will define the input vector to be $\vx = [s_t,1]\trp$, consisting of the signed stimulus intensity $s_t$ and a constant or ``bias'' term \cite{roy2021extracting}. This input interacts linearly with the weight vector \( \vw_t \), determining the probability that the animal selects the ``rightward'' choice on trial \( t \): 
\begin{equation} \label{eqn:glm}
p(y_t=1 \mid \vx_t, \vw_t) = \frac{1}{1 + e^{- \vx_t^\top \vw_t}}. 
\end{equation} 
The reward function is given by \( r_t = \indicator[y_t = z_t] \), which corresponds to a positive reward of \( 1 \) if the animal makes a correct decision (\( y_t=z_t \)) and a reward of \( 0 \) otherwise. 

\textbf{DNNs for learning rule inference.} 
Learning in this paradigm is captured by updates to the decision weights $\vw$. 
We first model the weight update function using a feedforward neural network (Fig.~\ref{fig:main_schematics}C):
\begin{equation}
\Delta \vw_t = f_{\theta}(\vw_t, \vx_t, y_t, r_t),
\end{equation}
where \( f_{\theta} \) is a feedforward network with two hidden layers and trainable parameters \( \theta \) (see Supp.~\ref{scn:sim_details} for architecture details). The next weight is given by:
\begin{equation}
\vw_{t+1} = \vw_t + \Delta \vw_t.
\end{equation}

Initial weights \( \vw_0 \) can be treated as trainable parameters, initialized via psychometric curve estimates (Supp.~\ref{scn:sim_details}). The inferred weights \( \vw_t \) parameterize the GLM response distribution (Eq.~\ref{eqn:glm}), and network parameters \( \theta \) are optimized to maximize the log-likelihood of observed choices:
\begin{equation}
\ell_t(\vw_t; \mathbf{x}_t) = y_t \log p(y_t=1 \mid \vx_t, \vw_t) + (1 - y_t) \log (1 - p(y_t=1 \mid \vx_t, \vw_t)),
\end{equation}
which corresponds to minimizing binary cross-entropy loss.

\textbf{RNNs for learning rule inference.} Since feedforward networks are memoryless, we also consider a recurrent parameterization using a GRU~\cite{cho2014learning} to encode history. The same inputs \( \{\vw_t, \vx_t, y_t, r_t\} \) are passed into a GRU \( g_{\theta} \) to produce a hidden state \( \mathbf{h}_t \), which is then passed to the feedforward network:
\begin{align}
\mathbf{h}_t &= g_{\theta}(\mathbf{h}_{t-1}, \vw_t, \vx_t, y_t, r_t), \\
\Delta \vw_t &= f_{\theta}(\mathbf{h}_t).
\end{align}
The model is trained using the same objective and procedure as above (see also Supp.~\ref{scn:sim_details}).

By combining a GLM for choice modeling with neural network–based learning rules, we balance interpretability and flexibility. We refer to the two models as \textbf{DNNGLM} and \textbf{RNNGLM}, depending on whether the learning rule is parameterized with a feedforward or recurrent network. Cross-validation, together with the implicit regularization properties of neural networks, helps mitigate overfitting; further details regarding our model and data processing can be found in Supp.~\ref{scn:sim_details}. The full procedure is outlined in Algorithm~\ref{alg:learning-rule-inference}. 

\begin{algorithm}[h!]
\caption{Learning Rule Inference from Trial-by-Trial Behavior}
\label{alg:learning-rule-inference}
\begin{algorithmic}[1]
\STATE \textbf{Inputs:} Trial sequence $\{s_t, y_t, r_t\}_{t=1}^{T}$
\FOR{trial $t = 1$ to $T$}
    \STATE Construct input vector: $u_t = [s_t, y_t, r_t, \vw_t]$
    \STATE Compute weight updates: $\Delta \vw_t \leftarrow \mathrm{NN}(u_t)$
    \STATE Update the inferred GLM weights: $\vw_{t+1} \leftarrow \vw_t + \Delta \vw_t$
    \STATE Predict choice probability: $P_{y_t} = \sigma(\vw_t^\top x_t)$
    \STATE Accumulate binary cross-entropy loss: $\mathcal{L}_t = -\left[y_t \log P_{y_t} + (1 - y_t) \log (1 - P_{y_t})\right]$
\ENDFOR
\STATE \textbf{Objective:} Minimize total loss $\mathcal{L} = \sum_{t=1}^T \mathcal{L}_t$ using gradient descent
\end{algorithmic} 
\end{algorithm} 

\section{Simulation results: recovery of ground-truth learning rules} 

\begin{figure}[h!]
    \centering
    \includegraphics[width=0.98\textwidth]{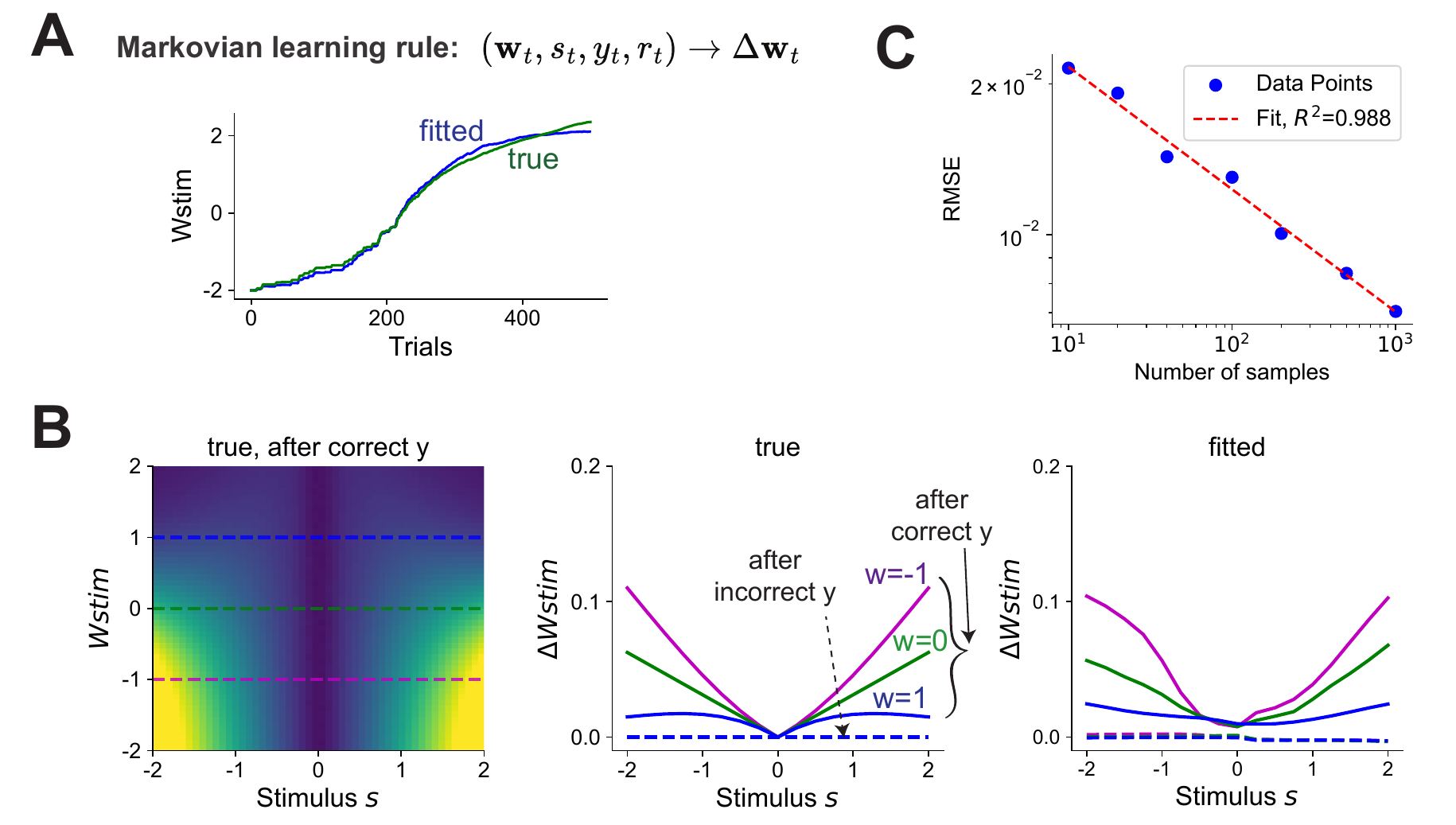}
    \caption{ \textbf{Recovering ground-truth learning rules in simulated data.} We simulated animal learning using a policy gradient method known as ``REINFORCE'' \cite{williams1992simple}.  (See Supp.\  Fig.~\ref{fig:supp_predMax} for an alternative ground-truth learning rule). This rule is Markovian in that the weight update $\Delta \vw_t$ depends only on current-trial variables. 
\textbf{(A)} Simulated trajectory for stimulus weight $\wstim$ for an example simulated mouse (green) and the inferred weights using the DNNGLM (blue).
\textbf{(B)} Left:  Heatmap showing the ground-truth stimulus weight change $\Delta \wstim$ following a correct choice, as a function of $\wstim$ and stimulus $s$. Horizontal dashed lines indicate $\wstim$ or $w$ slices shown in middle and right panels. Middle: slices showing stimulus weight change $\Delta \wstim$ as a function of the stimulus $s$, for different values of the true weight $\wstim$, after both correct (solid) and incorrect (dashed) decisions. (The REINFORCE algorithm exhibits no learning after incorrect trials for this setting of rule parameters). Right: corresponding slices through the weight update function inferred using the DNNGLM.  
Note that the model successfully captures key characteristics of the learning rule, such as the slowing of learning at higher weights, increased learning with stimulus amplitude, and the asymmetry in learning after correct and incorrect choices. 
(See Supp.\ Fig.~\ref{fig:supp_default} for analogous plots for bias parameter updates and Supp.\ Fig.~\ref{fig:supp_predMax} for a comparison to other learning rules.)
\textbf{(C)} Error in recovered learning rule as a function of dataset size, indicating that the DNNGLM converges to the true REINFORCE learning rule with increasing amounts of data.
} \label{fig:main_default}
\end{figure}

\begin{figure}[h!]
    \centering
    \includegraphics[width=0.98\textwidth]{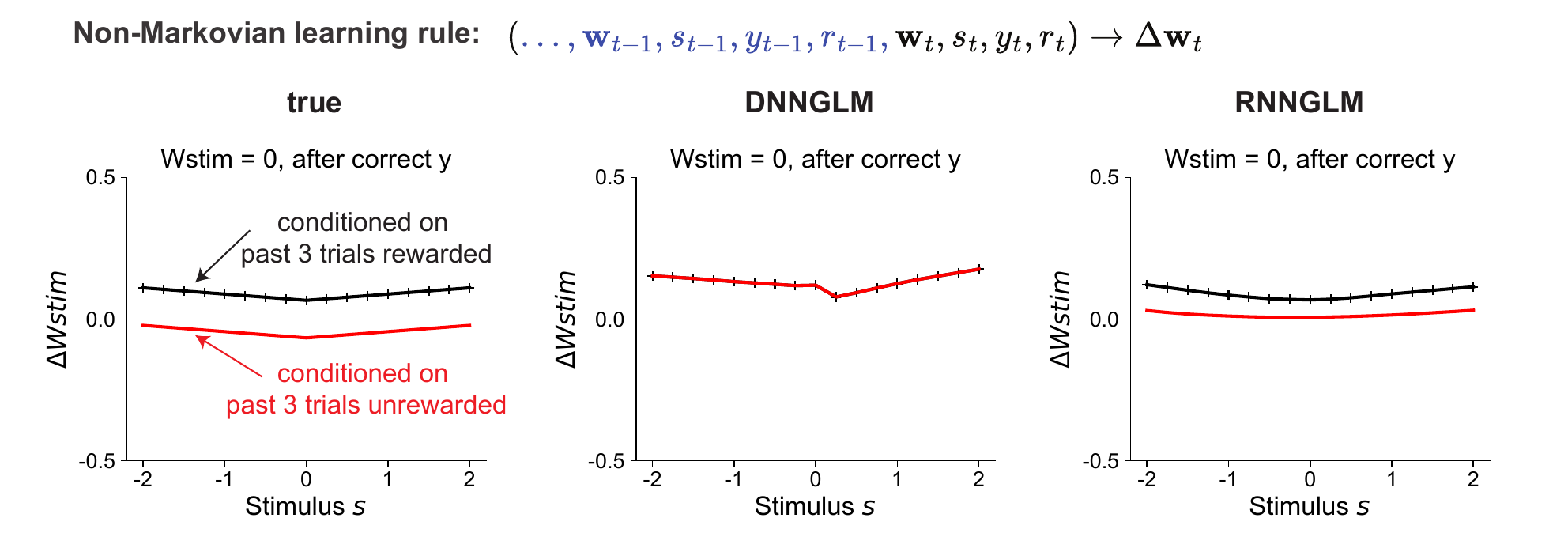}
    \caption{ \textbf{Simulated example with a non-Markovian learning rule. }
The {RNNGLM recovered the reward history dependence present in the ground truth: larger weight updates occur when the past $3$ trials were rewarded (black) versus unrewarded (red)}. Such differences were absent in DNNGLM, as it can only account for Markovian learning rules, where weight updates depend on current trial information. The ability of RNNGLM to capture these effects reflects its greater flexibility. 
Minor deviations from the ground truth are likely due to finite data effects (Fig.~\ref{fig:main_default}C). To keep the figure simple, we only plotted with the learning rule given stimulus weight $\wstim=0$ and after correct choices,  but we observed similar phenomena for other parameter settings.
    } \label{fig:main_nonMarkovian}
\end{figure}   

To begin our investigation, we first consider simulated data where the ground truth is known. We consider two types of learning rules: \textbf{Markovian} and \textbf{non-Markovian}. By Markovian learning rules, we mean that weight updates depend only on current-trial variables and are thus conditionally independent of stimuli, choices, and rewards on previous trials. 
In contrast, non-Markovian learning rules allow weight updates on the current trial to exhibit arbitrary dependencies on task history.  

For Markovian learning rules, we considered a classic policy gradient learning rule known as REINFORCE~\cite{williams1992simple}, which is defined by: 
\begin{align}
    \Delta \mathbf{\vw_t} &\propto r_t \nabla_{\vw} \log p(y_t \mid \vx_t, \mathbf{\vw_t}) \cr 
    &= r_t \epsilon_{y_t} (1 - p_{y_t}) x_t, 
\quad \epsilon_{y_t=R} = +1, \quad \epsilon_{y_t=L} = -1, 
\end{align} 
where $p_{y_t} := p(y_t \mid \vx_t, \mathbf{\vw_t})$, the probability of the choice $y_t$ given input $\vx_t$ under the animal's current policy. REINFORCE is a policy gradient method that uses Monte Carlo integration to evaluate the gradient of the policy with respect to reward (details in Supp.\ ~\ref{scn:reinf_details}). 

We first validated our framework on synthetic data where the ground truth learning rule is known (Fig.~\ref{fig:main_default}). When simulating animal learning under REINFORCE~\cite{williams1992simple}, our method (DNNGLM) recovered the weight trajectory and key properties of the learning rule, including dependencies on decision outcome, stimulus contrast, and current weight. Reconstruction accuracy---measured as the log base 10 of RMSE between the ground-truth $\Delta \vw$ and the model-predicted $\Delta \vw$ on the training data---also improves with increasing data (Fig.~\ref{fig:main_default}C).  

We then simulated a non-Markovian learning rule in which weight updates depended on trial history (Fig.~\ref{fig:main_nonMarkovian}). Specifically, we consider a modified REINFORCE rule with an eligibility-trace–like factor derived from a non-Markovian task setting (see Supp.~\ref{scn:etrace_details} for details):
\begin{equation}
    \Delta \vw_t \propto r_t \sum_s \epsilon_{y_{t-s}} (1 - p_{y_{t-s}}) \vx_{t-s}.
\end{equation}
A key property of this rule is that a correct and rewarded past trial contributes a nonnegative term to the sum --- since \( \epsilon_{y_{t-s}} \) and \( \vx_{t-s} \) share the same sign and the remaining terms are nonnegative --- while an incorrect trial contributes a nonpositive term. Thus, conditioning on several past trials being rewarded (versus unrewarded) leads to larger weight updates under the ground truth learning rule. This history-dependent effect is successfully captured by the fitted RNNGLM (Fig.~\ref{fig:main_nonMarkovian}). 

In this case, the DNNGLM model fails to recover the correct learning dynamics; RNNGLM recovers the reward-history dependence in the learning rule.  This demonstrates the need for recurrent architectures in modeling non-Markovian, history-dependent learning processes. In principle, the weights \( \vw \) could carry some history even in DNNs, since \( \vw \leftarrow \vw + \Delta \vw \) integrates past updates. However, the learned update function \( \Delta \vw \) itself does not explicitly depend on past trial information and thus cannot capture history-dependent learning dynamics (see Fig.~\ref{fig:main_nonMarkovian}). While expanding the DNN input to include historical information is possible, this would require manual feature engineering and reduce the model’s ability to automatically determine which past inputs are relevant. 

\begin{figure}[h!]
    \centering
    \includegraphics[width=0.98\textwidth]{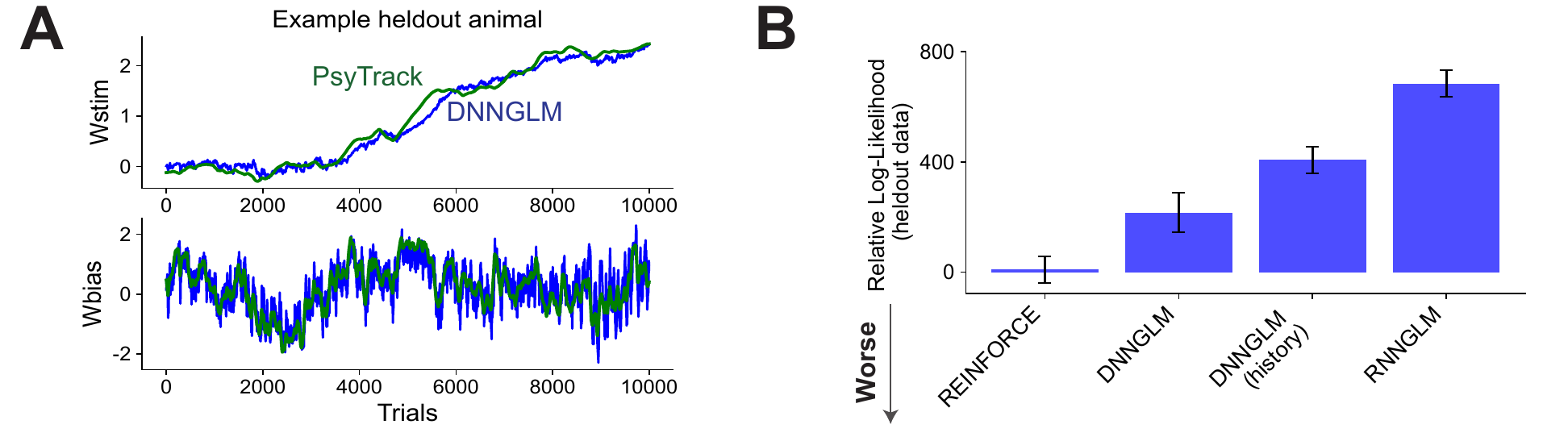}
    \caption{\textbf{Application to IBL mouse learning
    data.} 
\textbf{(A)} Predicted stimulus (top) and bias (bottom) weight trajectories for a held-out animal --- using DNNGLM trained on other animals and applied to this animal’s stimulus and choice sequence --- plotted along the weight trajectories inferred by PsyTrack \cite{roy2018efficient}, which is a learning-agnostic and reliable method for tracking psychometric weights from behavioral time series. (Supp.\ Fig.~\ref{fig:supp_psyDNN} shows additional weight trajectories.) Similar traces were observed for RNNGLM as well. 
\textbf{(B)} Our methods  (DNNGLM and RNNGLM) achieved significantly higher test log-likelihood (LL) on held-out data. Here we plot LL relative to the REINFORCE model. Notably, the RNNGLM model also out-performed the DNNGLM extended to include the previous trial input (``DNNGLM- history''). Error bars reflect standard deviation across cross-validation seeds. 
    } \label{fig:main_ibl} 
\end{figure}

We also add more results in the Supp.\ Material demonstrating that our main findings are robust across a variety of conditions: (i) a learning rule without the decision outcome asymmetry (Supp.\  Fig.~\ref{fig:supp_predMax}); (ii) recovery of an in-between learning rule when mixing update functions across individuals (Supp.\  Fig.~\ref{fig:supp_mixed_rates}); (iii) different initial stimulus weights (Supp.\  Fig.~\ref{fig:supp_unifW0}); (iv) longer trial lengths (Supp.\  Fig.~\ref{fig:supp_T8000}). While the initial policy weights are typically unknown, we estimated them from the psychometric curve (Supp.~\ref{scn:sim_details}) and demonstrated recovery in Supplementary Fig.~\ref{fig:supp_T8000}. For faster simulations in the main text, and to allow multiple repetitions for robustness testing, we used shorter trial sequences. These shorter sequences make it difficult to estimate the initial weight due to fewer samples at the beginning, so we initialized with a known initial weight in this case. Robustness tests showing recovery with incorrect $W_0$ are provided in Supplementary Table~\ref{tab:w0_sensitivity}. We also show sensitivity to the initial weight estimate (Supp.\  Table~\ref{tab:w0_rmses}) and degradation of reconstruction under added noise (Supp.\  Table~\ref{tab:rmse_noise}). 

\section{Application to mouse learning data} 

\begin{figure}[h!] 
    \centering
    \includegraphics[width=0.98\textwidth]{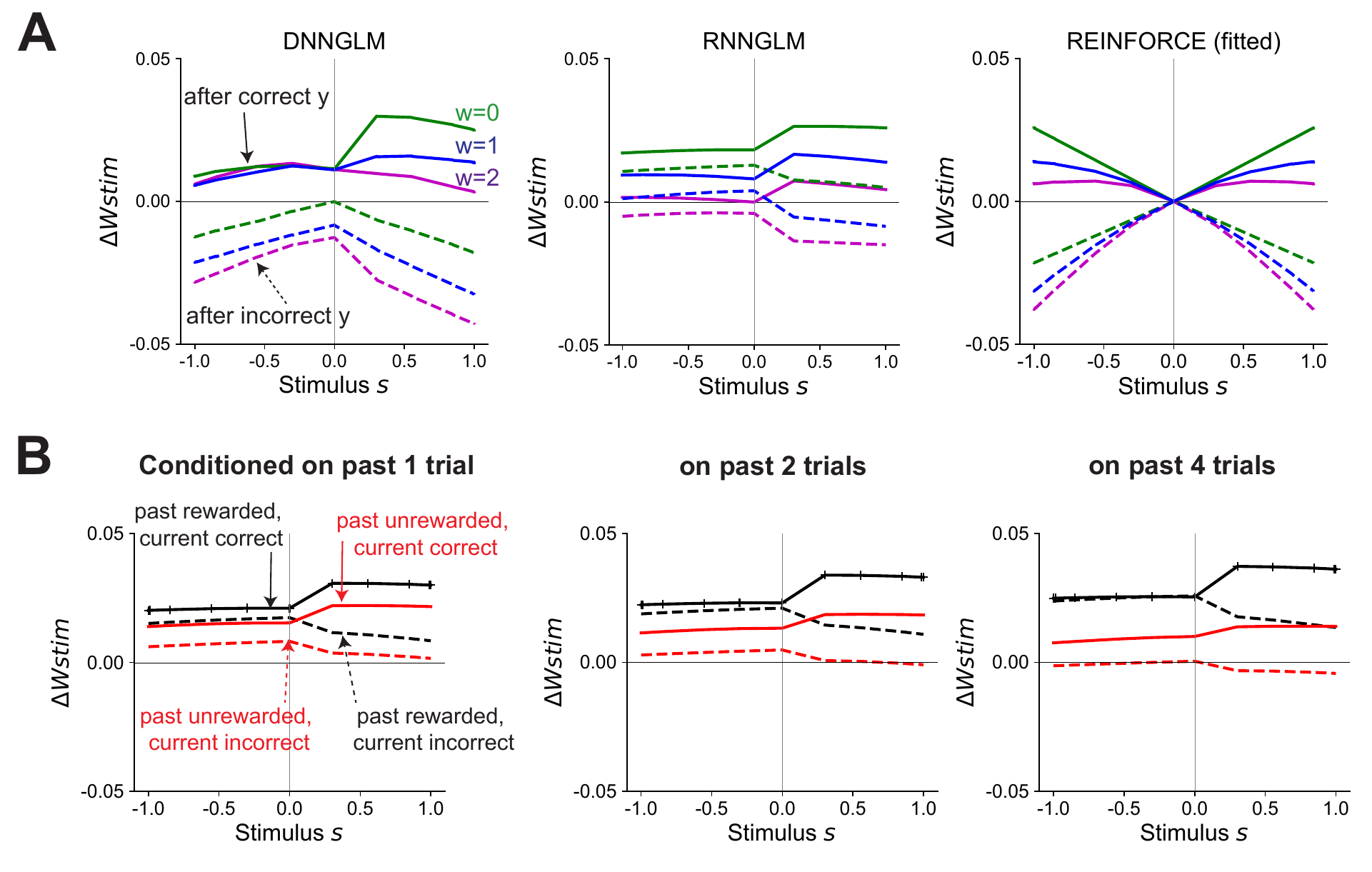}
    \caption{ \textbf{Properties of inferred learning rules.} 
\textbf{(A)} For the dataset in Fig.~\ref{fig:main_ibl}, DNNGLM reveals negative weight updates following errors (coined as ‘negative baseline’ in~\cite{geadah2025inferring}, which was also observed in our flexible framework). Bias weight plot is provided in Supp.\ Fig.~\ref{fig:supp_IBL_db}; note $\vw$ is fixed at positive weights because the training data mainly involved positive $\vw$. We follow the same plotting convention in Fig.~\ref{fig:main_default}; since RNNGLM depends on additional historical variables, we plot the mean \( \Delta \wstim \) averaged across these historical dimensions. 
\textbf{(B)} RNNGLM suggests non-Markovianity (history dependence) in learning: larger weight updates are observed when all past trials were rewarded (black) versus unrewarded (red). This gap widens as we condition on history beyond just the previous trial, \textbf{suggesting that the history dependency extends beyond the most recent past trial}. We note that plotting up to four past trials was intended only as an illustrative example, not a fundamental limit of RNNGLM; as shown in Supp.\ Fig.~\ref{fig:supp_toffset}, the model suggests longer history dependencies. To keep the figure simple, we only plotted with stimulus $w$ fixed at 0, but similar trends are observed elsewhere. 
    } \label{fig:main_ibl2} 
\end{figure} 

For real data, we used the publicly available International Brain Laboratory (IBL) dataset~\cite{international2020standardized}, following the same preprocessing as in~\cite{ashwood2020inferring}. We first fit the learning rule on a pool of training animals. We then took the stimulus and choice sequence from a held-out test animal and applied the fitted DNNGLM or RNNGLM to predict the evolution of its stimulus and bias weights. We compared the predicted weight trajectories to those inferred by PsyTrack~\cite{roy2018efficient}, a learning-agnostic and reliable method for tracking psychometric weights from behavior, and observed close agreement (Fig.~\ref{fig:main_ibl}A). Furthermore, using the predicted weights to compute the choice log-likelihood on held-out animals, both DNNGLM and RNNGLM outperformed REINFORCE-based models (Fig.~\ref{fig:main_ibl}B and Table~\ref{tab:heldout-LL-results}). We also assessed temporal generalization on future held-out data and observed similar trends (Supp.~\ref{tab:future_LL}). Notably, RNNGLM also outperformed history-augmented DNNGLM, in which previous-trial information was added as an additional regressor to the GLM. 

Moreover, our methods revealed empirical features of mouse learning that deviate from classical learning rules (Fig.~\ref{fig:main_ibl2}A). These include the presence of a ``negative baseline'' (coined in~\cite{geadah2025inferring} but also observed in our flexible framework), where weight updates following incorrect trials actually \textit{decreased} task accuracy --- even though a positive update would aid learning in this particular task. We also observed asymmetries in update magnitude based on decision correctness, with distinct updates following correct (solid lines) versus incorrect (dashed lines) choices, as well as asymmetries across stimulus sides (i.e., different update functions for left vs.\ right choices). Notably, such side asymmetries have also been reported in prior work~\cite{liebana2023striatal}.

In addition, the fitted RNNGLM --- achieving the best held-out log-likelihood --- showed higher updates following sequences of rewarded trials (Fig.~\ref{fig:main_ibl2}B), suggesting sensitivity to reward history spanning multiple trials (see also Supp.~Fig.~\ref{fig:supp_mixed_rates}B). Together, these findings highlight the framework’s ability to improve predictive accuracy while uncovering mechanistic insights into animal learning. 

Because inferring flexible learning rules from binary choices is an under-constrained inverse problem, multiple parameterizations can fit the data similarly well. We therefore repeated inference for each model (DNNGLM, RNNGLM) across multiple random seeds, holding data splits and hyperparameters fixed. Across seeds, log-likelihoods were similar, while the recovered update functions showed modest variation. To visualize this, we now plot mean and standard deviation across weight updates across seeds (Supp.\ Fig.~\ref{fig:supp_shaded}). Despite some variability, \textbf{the key characteristics are consistent across seeds}—a negative baseline, side asymmetry, and multi-trial (non-Markovian) dependence on reward/stimulus history. Moreover, the elimination of these features suggests they are functionally meaningful: enforcing a non-negativity constraint on the baseline reduces held-out predictive performance (Table~\ref{tab:heldout-LL-results}). Additionally, to assess fairness, we matched parameter counts between DNNGLM and RNNGLM, and observed that RNNGLM still outperformed ($p=5.53e-4$), indicating the gap is not due to just model size.

\begin{table}[t]
\centering
  \begin{tabular}{lcc}
    \toprule
    \textbf{Comparison} & \textbf{Heldout LL} & \textbf{$p$-val} \\
    \midrule
    REINFORCE vs. \textbf{DNNGLM}                   & $-60576$ vs. $\mathbf{-60360}$       & $1.57\text{e}{-2}$ \\
    DNNGLM vs. \textbf{RNNGLM}                      & $-60360$ vs. $\mathbf{-59892}$       & $1.79\text{e}{-8}$ \\
    \textbf{REINFORCE} vs. nonnegative base         & $\mathbf{-60576}$ vs. $-68532$       & $1.25\text{e}{-5}$ \\
    \midrule
    tinyRNN vs. \textbf{RNNGLM}                     & $-60564$ vs. $\mathbf{-59892}$       & $2.12\text{e}{-3}$ \\
    REINFORCE (history) vs. \textbf{RNNGLM}         & $-60504$ vs. $\mathbf{-59892}$       & $7.73\text{e}{-7}$ \\
    DNNGLM (history) vs. \textbf{RNNGLM}            & $-60168$ vs. $\mathbf{-59892}$       & $3.84\text{e}{-6}$ \\
    \bottomrule
\end{tabular}
  \caption{
Animal-level held-out log-likelihoods (higher is better) for Fig.~\ref{fig:main_ibl}B, except that we report total log-likelihood (summed across animals) instead of relative log-likelihood. 
We show $p$-values from paired $t$-tests across mice, comparing different learning rule models on the IBL dataset. Each entry shows two values (one per model in the comparison), reported as mean ± standard error across cross-validation seeds. For the interpretable RNN baseline --- which predicts behavior directly rather than parameterizing weights or weight updates --- we used tinyRNN~\cite{ji2023automatic} with 8 GRUs, more than the 1–4 GRUs used in their original paper, which performed even worse in our tests. Results for held-out future data also show consistent trends (Supp.\  Table~\ref{tab:future_LL}). 
}
  \label{tab:heldout-LL-results}
\end{table}

\section{Related Work} 

Previous studies have shown that artificial neural networks provide a more accurate description of animal learning behavior than classical learning models 
\cite{richards2019deep, yang2020artificial, yamins2014performance,banino2018vector, cueva2018emergence,whittington2020tolman,tyulmankov2022meta,mante2013context,pagan2025individual,wang2018prefrontal,fascianelli2022neural,jensen2024recurrent,peterson2021using,song2021using,jaffe2023modelling,kuperwajs2023using,rmus2024artificial,ger2024harnessing}. Several recent works have taken steps to enhance the interpretability of inferred learning strategies --- for example, by regularizing the latent space or parameterizing in low-dimensional or human-interpretable forms~\cite{ji2023automatic,miller2023cognitive,eckstein2023predictive,castro2025discovering}, but they have not addressed \textit{de novo} learning, i.e., learning to perform a new sensory-motor task behavior structure (input–output mappings) from scratch.

Methods for inferring learning rules during the acquisition of new task structures remain sparse. Those that do exist typically assume a predefined functional form~\cite{Lak2020elife,ashwood2020inferring,liebana2023striatal,geadah2025inferring}, limiting their ability to capture learning strategies that differ from hand-designed models. In addition, many are memoryless or 'Markovian', assuming that weight updates depend only on the task variables encountered on the current trial; this stands in contrast to biological learning, where history plays a prominent role --- such as through eligibility traces~\cite{Magee2020,Gerstner2018} or the influence of trial history on decision-making~\cite{roy2021extracting,urai2017pupil,morcos2016history,akrami2018posterior}. In this work, we focus on inferring learning rules from behavioral data,
motivated by the broad availability of behavioral datasets and the potential for large-scale inference without requiring simultaneous neural recordings. 
In parallel, many studies have proposed approaches to study synaptic-level plasticity rules~\cite{lim2015inferring,portes2022distinguishing,kepple2021curriculum,nayebi2020identifying,mehta2024model,confavreux2020meta,confavreux2023meta,ramesh2023indistinguishable,bell2024discovering,liu2021cell,liu2022biologically,lillicrap2016random,hinton2022forward}. While some of these approaches can also be applied to behavioral data (e.g.,~\cite{mehta2024model}), their focus has largely been on synaptic-level inference and often requires neural activity recordings or handcrafted functional bases for interpretation. Our method instead emphasizes flexible inference directly from behavior, improving weight recovery via psychometric-curve initialization and extending flexibility with recurrent networks to capture multi-trial history beyond bandit tasks. 

Beyond neuroscience, our work is also related to the meta-learning literature. A prominent line of work learns optimizers or update rules via gradient-based meta-learning~\cite{andrychowicz2016learning}, with extensions to reinforcement learning~\cite{duan2016rl2,wang2018prefrontal}; several computational neuroscience studies likewise meta-learn synaptic plasticity rules within interpretable frameworks~\cite{miconi2018differentiable,confavreux2020meta,tyulmankov2022meta,shervani2023meta}. In contrast to meta-learning approaches that optimize agents to learn well across tasks, we address a fundamentally different goal: inferring, from behavior alone and in \textit{de novo} tasks, the learning rule that a biological agent used to learn from scratch. This poses distinct challenges: (i) animals often exhibit suboptimal or biased strategies that violate normative reward-maximization assumptions~\cite{akrami2018posterior,kastner2022spatial}, enlarging the solution space and introducing history-dependence and other idiosyncrasies; (ii) updates must integrate external stimuli—not just reward histories—going beyond classic bandit settings; and (iii) many prior approaches fix feature representations, use static policy forms, or rely on hand-designed rules that do not adapt during training. These factors call for flexible yet interpretable models with minimal structural priors; under these conditions, prior baselines underperform relative to ours (Table~\ref{tab:heldout-LL-results}). Our approach thus helps bridge structured neuroscience models with black-box meta-learning in this \textit{de novo} inference setting.

\section{Discussion} 

Understanding how animals learn from experience—particularly in \textit{de novo} settings where input–output mappings must be acquired from scratch—remains a central challenge. Prior work has largely relied on predefined parametric models or focused on simpler settings like bandit tasks that do not require learning a new task structure from scratch. Here we instead infer, from behavior alone, the trial-by-trial learning rule used by the animal, confronting three challenges highlighted in Related Work: suboptimal or biased updates, the need to integrate external stimuli and multi-trial history, and the limits of fixed rule forms or static policy features. We demonstrate that our framework recovers ground-truth rules in simulation, extends to non-Markovian dynamics via recurrent architectures, improves held-out prediction accuracy on real behavioral data, and yields analyzable insights into animal learning. Together, these results support a data-driven framework for uncovering learning processes. 

To our knowledge, this is the first study to infer the learning rule function in \textit{de novo} learning settings without assuming a specific model class. While neural network–based approaches have been proposed in non-\textit{de novo} settings such as bandit tasks~\cite{dezfouli2019disentangled,eckstein2023predictive,miller2023cognitive,ji2023automatic,castro2025discovering}, structured sensory inputs and the need to learn new input–output mappings introduce additional challenges. Prior work has also modeled behavior by having RNNs directly output cognitive states or choices, but such models often act as black boxes that are difficult to interpret~\cite{ji2023automatic,miller2023cognitive}. Although recent studies have sought to balance flexibility and interpretability by using RNNs to represent classical model parameters~\cite{ger2024harnessing}, we find that directly modeling parameter updates $\Delta \vw$ leads to better ground truth reconstruction (Supp.~Fig.~\ref{fig:supp_RNNW}). This advantage likely stems from the implicit regularization of neural networks, which biases learning dynamics toward smoother functions first in the feature learning regime~\cite{jacot2018neural,chizat2019lazy,flesch2021rich,george2022lazy,braun2022exact,pezeshki2021gradient,baratin2021implicit,rahaman2019spectral,cao2019towards,atanasov2021neural,canatar2021spectral,emami2021implicit,wilson2025deep}. That said, it remains unclear whether the inductive bias of our setup generalizes across various learning systems --- an open question we highlight as a future direction (see below). One promising result supporting our approach is that, in a sense, implicitly regularizing weight updates aligns with the philosophy behind PsyTrack, which also acts at the level of weight changes~\cite{roy2018efficient}. 

\textbf{Broader outlook for neuroscience and AI}. By inferring learning rules directly from behavior without assuming a fixed model structure, our approach offers a data-driven framework for understanding how animals learn \textit{de novo}. This has several implications. First, interpreting the inferred learning rules could offer insights into learning behavior (e.g., suboptimal or history-dependent patterns), which may point to additional variables—such as trial history, internal states, or reward timing—that are worth tracking in future behavioral learning studies. In turn, more accurate behavioral learning models could inform applications in health and conservation—e.g., anticipating how animals respond to environmental changes or interventions~\cite{gnanasekar2022rodent,shaw2023human,manduca2023learning,hart2011behavioural,wijeyakulasuriya2020machine,hou2020identification,ditria2020automating,pillai2023deep,nilsson2020simple}. This also lays a foundation for more challenging settings with richer task structures and scaling to more expressive architectures. Second, \textit{de novo} learning is ubiquitous in experimental neuroscience, where animals are often trained from scratch on novel tasks, yet remains understudied. By revealing structure in how animals learn during this phase, our method can inform experimental design and training protocols, supporting more efficient task training~\cite{ashwood2020inferring,geadah2025inferring}. 

From an AI perspective, our work contributes to the growing interest in biologically-aligned learning systems~\cite{sucholutsky2023getting,zador2023catalyzing}. By uncovering suboptimality and history dependence in animal learning, we reveal deviations from standard artificial agents that could inform the design of more human- or animal-aligned AI. Relatedly, our approach may support the development of behavioral digital twins—AI models that mimic individual learning processes for simulation, personalization, or real-time prediction~\cite{katsoulakis2024digital}. Finally, by leveraging the implicit regularization of neural networks to infer learning rules, our work raises new questions about how such inductive biases interact with neural network–based system identification (see below).

\textbf{Limitations and future directions.}  
\textit{Pooling across animals.} Neural network–based approaches are often data hungry, and reconstruction accuracy improves with more data (Fig.~\ref{fig:main_default}C). To address this, we pool across animals rather than fitting individual subjects. While pooling is common in neuroscience~\cite{ashwood2022dynamic,ashwood2022mice,bolkan2022opponent}, it can obscure individual variability and inherit known limitations~\cite{moore2022slow}. Nonetheless, learning rule inference from behavior alone is underconstrained in low-data regimes, and pooling offers a practical mitigation strategy. This population-level approach aligns with broader trends in neuroscience and AI toward large-scale behavioral datasets. 

\textit{Static vs. dynamic learning rules.} Our current model assumes a static weight update function. While RNNs may implicitly capture temporal changes—perhaps explaining why fitting separate RNNGLMs to the first and second halves of training did not yield improvement—future work could explicitly model dynamic learning rules to better disentangle non-Markovian dependencies from nonstationarity. An important direction is to investigate whether the observed suboptimal negative baseline arises from specific cognitive mechanisms --- such as choice perseverance or forgetting~\cite{davis2017biology} --- or from model mismatch. More broadly, suboptimal learning is common in the literature~\cite{molano2023recurrent,ma2024vast}, especially in pathological populations~\cite{dezfouli2019models,kumar2025neurocomputational}, where applying our approach could be interesting.

\textit{Stochasticity.} Our current learning model is deterministic. Capturing stochasticity in learning would be a valuable extension.

\textit{Generalization across datasets, tasks, and architectures.} We evaluated our method using the same preprocessed IBL dataset as~\cite{ashwood2020inferring}. This focus on a relatively simple task was intentional: it mirrors widely used \textit{de novo} learning paradigms in laboratory neuroscience, where animals are introduced to novel tasks with minimal prior structure. Such simplified tasks capture key hallmarks of biological learning—history dependence, biases, and suboptimal strategies—while providing a tractable platform for reverse-engineering learning dynamics. In this sense, our work should be viewed as a stepping stone toward broader generalization: establishing validity in a controlled setting before extending to richer tasks and datasets. Indeed, to demonstrate potential scalability, we tested higher-dimensional inputs and found that our approach continues to recover core qualitative features of learning rules, with performance improving as data availability increases (Supp.\ Table~\ref{tab:input-dim}). These results indicate that our framework can extend beyond the simple tasks considered here. 

Beyond extending to diverse tasks and datasets, another future avenue would be to investigate how different architectures—and their associated inductive biases—affect inference accuracy, particularly under data limitations. Our framework currently fixes the decision model (e.g., standard models such as GLMs), which aids interpretability and follows common neuroscience practice but limits expressivity. In fact, for multiplicative decision processes, this assumption leads to a significant performance degradation (Supp.\ Table~\ref{tab:multiplicative-decision}). We view this modular approach—using a standard decision model while focusing on trial-by-trial weight updates—as a principled first step, consistent with strategies in neuroscience and machine learning where components are studied separately before joint inference. Extending the framework to jointly infer both the decision model and the learning rule is an important direction for future work. 

Overall, we introduce a flexible framework for inferring learning rules directly from behavior, without assuming a predefined form. By capturing non-Markovian dynamics and uncovering insights into \textit{de novo} learning, we hope this approach opens new avenues for studying how animals learn in rich, structured environments.

\section*{Acknowledgement}

The authors thank the International Brain Laboratory for data availability, Orren Karniol-Tambour for feedback on the manuscript, and Alex Riordan for helpful discussions.  
YHL was partially supported by the Fonds de recherche du Québec – Nature et technologies (FRQNT). 
VG was supported the Porter Ogden Jacobus Fellowship at Princeton University, and by doctoral scholarships from the Natural Sciences and Engineering Research Council of Canada (NSERC) and the Fonds de recherche du Québec – Nature et technologies (FRQNT). 
JWP was supported by grants from the Simons Collaboration on the Global Brain (SCGB AWD543027), the NIH BRAIN initiative (9R01DA056404-04), an NIH R01 (NIH 1R01EY033064), and a U19 NIH-NINDS BRAIN Initiative Award (U19NS104648).

\bibliographystyle{unsrt}
\bibliography{ref_main} 


\newpage 
\section*{NeurIPS Paper Checklist}

\begin{enumerate}

\item {\bf Claims}
    \item[] Question: Do the main claims made in the abstract and introduction accurately reflect the paper's contributions and scope?
    \item[] Answer: \answerYes{} 
    \item[] Justification: To make this easier for the readers, we have referred to the pertinent figures and sections under "Main contributions" in Introduction. 

\item {\bf Limitations}
    \item[] Question: Does the paper discuss the limitations of the work performed by the authors?
    \item[] Answer: \answerYes{} 
    \item[] Justification: Details on limitations and future work are discussed in the 'Limitations and future works' subsection in Discussion. 

\item {\bf Theory assumptions and proofs}
    \item[] Question: For each theoretical result, does the paper provide the full set of assumptions and a complete (and correct) proof?
    \item[] Answer: \answerYes{} 
    \item[] Justification: We provide detailed assumption and proof for Proposition~\ref{prop:SE} in Supp.\ ~\ref{scn:theory}. 

    \item {\bf Experimental result reproducibility}
    \item[] Question: Does the paper fully disclose all the information needed to reproduce the main experimental results of the paper to the extent that it affects the main claims and/or conclusions of the paper (regardless of whether the code and data are provided or not)?
    \item[] Answer: \answerYes{} 
    \item[] Justification: Training details are provided in Supp.\ ~\ref{scn:sim_details}. Moreover, our code is available \url{https://github.com/Helena-Yuhan-Liu/InferLearningANNGLM}

\item {\bf Open access to data and code}
    \item[] Question: Does the paper provide open access to the data and code, with sufficient instructions to faithfully reproduce the main experimental results, as described in supplemental material?
    \item[] Answer: \answerYes{} 
    \item[] Justification: We use IBL datasets, which are publicly available. Our code is available \url{https://github.com/Helena-Yuhan-Liu/InferLearningANNGLM}.

\item {\bf Experimental setting/details}
    \item[] Question: Does the paper specify all the training and test details (e.g., data splits, hyperparameters, how they were chosen, type of optimizer, etc.) necessary to understand the results?
    \item[] Answer: \answerYes{} 
    \item[] Justification: Details are provided in Supp.\ ~\ref{scn:sim_details}. 

\item {\bf Experiment statistical significance}
    \item[] Question: Does the paper report error bars suitably and correctly defined or other appropriate information about the statistical significance of the experiments?
    \item[] Answer: \answerYes{} 
    \item[] Justification: We have included this information in all applicable figures and tables. In particular, $p$-values for log-likelihood comparisons are provided in Table~\ref{tab:heldout-LL-results}, with details of the $t$-test and cross-validation procedure described in Supp.\ ~\ref{scn:sim_details}.

\item {\bf Experiments compute resources}
    \item[] Question: For each experiment, does the paper provide sufficient information on the computer resources (type of compute workers, memory, time of execution) needed to reproduce the experiments?
    \item[] Answer: \answerYes{} 
    \item[] Justification: Information pertaining to computing resources and simulation time can be found in Supp.\ ~\ref{scn:sim_details}. 
    
\item {\bf Code of ethics}
    \item[] Question: Does the research conducted in the paper conform, in every respect, with the NeurIPS Code of Ethics \url{https://neurips.cc/public/EthicsGuidelines}?
    \item[] Answer: \answerYes{} 
    \item[] Justification: We have carefully read the NeurIPS Code of Ethics and attest that the research conforms.

\item {\bf Broader impacts}
    \item[] Question: Does the paper discuss both potential positive societal impacts and negative societal impacts of the work performed?
    \item[] Answer: \answerNA{} 
    \item[] Justification: This research advances our understanding of animal learning mechanisms and contributes to the development of interpretable models for behavior. We do not anticipate any immediate ethical or societal impacts. However, over time, these findings could influence related fields such as neuroscience and machine learning, which may indirectly affect society depending on how these technologies are applied.
    
\item {\bf Safeguards}
    \item[] Question: Does the paper describe safeguards that have been put in place for responsible release of data or models that have a high risk for misuse (e.g., pretrained language models, image generators, or scraped datasets)?
    \item[] Answer: \answerNA{} 
    \item[] Justification: No specific safeguards were implemented, since this work is focused on basic research to better understand animal learning, as explained above. The model is not intended for deployment or real-world decision-making. As such, no foreseeable misuse or ethical risks require mitigation at this stage.

\item {\bf Licenses for existing assets}
    \item[] Question: Are the creators or original owners of assets (e.g., code, data, models), used in the paper, properly credited and are the license and terms of use explicitly mentioned and properly respected?
    \item[] Answer: \answerYes{} 
    \item[] Justification: Please see Supp.\ ~\ref{scn:sim_details}. 

\item {\bf New assets}
    \item[] Question: Are new assets introduced in the paper well documented and is the documentation provided alongside the assets?
    \item[] Answer: \answerNA{} 
    \item[] Justification: The paper does not release new assets. 

\item {\bf Crowdsourcing and research with human subjects}
    \item[] Question: For crowdsourcing experiments and research with human subjects, does the paper include the full text of instructions given to participants and screenshots, if applicable, as well as details about compensation (if any)? 
    \item[] Answer: \answerNA{} 
    \item[] Justification: The paper does not involve crowdsourcing nor research with human subjects.

\item {\bf Institutional review board (IRB) approvals or equivalent for research with human subjects}
    \item[] Question: Does the paper describe potential risks incurred by study participants, whether such risks were disclosed to the subjects, and whether Institutional Review Board (IRB) approvals (or an equivalent approval/review based on the requirements of your country or institution) were obtained?
    \item[] Answer: \answerNA{} 
    \item[] Justification: The paper does not involve crowdsourcing nor research with human subjects.

\item {\bf Declaration of LLM usage}
    \item[] Question: Does the paper describe the usage of LLMs if it is an important, original, or non-standard component of the core methods in this research? Note that if the LLM is used only for writing, editing, or formatting purposes and does not impact the core methodology, scientific rigorousness, or originality of the research, declaration is not required.
    \item[] Answer: \answerNA{} 
    \item[] Justification: LLMs were used solely for writing, editing, and formatting assistance in this work. We also briefly used Deep Research to assess the coverage of our related work discussion and found that our initial literature review was already comprehensive. 

\end{enumerate}

\newpage 


\appendix

\section{Theoretical justifications} \label{scn:theory}

To gain theoretical insights into why this approach could work, we consider two key questions: (1) Do neural networks have the capacity to approximate the learning rule, i.e., the weight update function? (2) Since weight updates are not directly observable, can they be identified from behavioral data? Question (1), assuming well-behaved weight update functions, is addressed by universal approximation theorems~\cite{cybenko1989approximation,hornik1991approximation,pinkus1999approximation,hassoun1995fundamentals,haykin1998neural,lu2017expressive,funahashi1993approximation}. For question (2), we start by considering a very simple idealized setting in the proposition below and leave comprehensive theoretical investigations into identifiability for future work.  

\begin{prop} \label{prop:SE} 
For simplicity, let \( \vw_t \in \RR \) for \( t \in \{0, 1\} \), and assume \( y_t \sim \mathrm{Bernoulli}(P_{y_t}) \), where
\[
P_{y_t} := \mathbb{P}[y_t = 1 \mid \vw_t, x_t] = \sigma(\vw_t x_t), \quad \text{with } \sigma(z) = \frac{1}{1 + e^{-z}} \text{ denoting the sigmoid function.}
\]
We further make the following simplifying assumptions: (1) \(\Delta \vw_t = \vw_{t+1} - \vw_t\) depends only on \((\vw_t, x_t, y_t)\). (2) For each \(t\), inputs \(x_t^{(i)} \in \{-1, 1\}\) are drawn i.i.d. across samples from the uniform distribution over this set, i.e., \( \mathbb{P}(x_t = -1) = \mathbb{P}(x_t = 1) = \frac{1}{2} \). (3) The same initial weight $\vw_0$ is used across $N$ independent repetitions. (4) For all sample \(i\), we assume \(P_{y_t}^{(i)} \in [\epsilon, 1 - \epsilon]\) for some arbitrary \(0 < \epsilon < 1\). (5) The prior on \(\vw_t\) is smooth, strictly positive, and does not concentrate near the boundary of the parameter space. Then, as $N$ approaches infinity, the standard error (SE) of \(\Delta \vw_0 = w_1 - w_0\) across samples approaches $0$.
\end{prop} 

\begin{proof}

The log-likelihood for a single observation at time $t$ is:
\[
\ell_t(\vw_t) = y_t \log P_{y_t} + (1 - y_t) \log (1 - P_{y_t}). 
\]

By taking the second derivative, we arrive at:
\begin{align*}
\frac{\partial^2 \ell_t(\vw_t)}{\partial \vw_t^2} &= -P_{y_t} (1 - P_{y_t}) x_t^2 \cr &= -P_{y_t} (1 - P_{y_t}),  
\end{align*}
where $x_t^2$ is dropped due to the assumption that $x \in \{-1, 1\}$. 

Summing over \(N\) independent observations, the Fisher information for \(\vw_t\) is:
\[
I(\vw_t) = \sum_{i=1}^N P^{(i)}_{y_t}(1-P^{(i)}_{y_t}).  
\]

By the Bernstein-von Mises (BvM) theorem (see \cite{van2000asymptotic}), the posterior distribution of $\vw_t$ given the data is asymptotically normal:
\[
\pi(\vw_t \mid \text{data}) \xrightarrow{d} \mathcal{N}\left(\hat{w}_t, \frac{1}{I(\vw_t)} \right),
\]
where $\hat{w}_t$ is the MLE of $\vw_t$.

Thus, by BvM theorem, in the limit as \(N \to \infty\), the posterior variance is asymptotically equal to:
\[
\text{Var}(\vw_t) = \frac{1}{\sum_{i=1}^N P^{(i)}_{y_t}(1 - P^{(i)}_{y_t})}.
\] 

and applying assumption (4):
\[
\text{Var}(\vw_t) \leq \frac{1}{N \epsilon (1 - \epsilon)}, 
\]
and 
\[
\text{SE}(\vw_t) \leq \sqrt{\frac{1}{N \epsilon (1 - \epsilon) }}.
\]

Since $\Delta \vw_0 = w_1 - w_0$, the variance of $\Delta \vw_0$ can be expressed as:
\[
\text{Var}(\Delta \vw_0) = \text{Var}(w_1) + \text{Var}(w_0) - 2 \cdot \text{Cov}(w_1, w_0).
\]

Using the Cauchy-Schwarz inequality, the covariance term is bounded:
\[
|\text{Cov}(w_1, w_0)| \leq \sqrt{\text{Var}(w_1) \cdot \text{Var}(w_0)}.
\]

Thus, we have:
\[
\text{Var}(\Delta \vw_0) \leq \text{Var}(w_1) + \text{Var}(w_0) + 2 \sqrt{\text{Var}(w_1) \cdot \text{Var}(w_0)}.
\]

Substituting the variance bounds:
\[
\text{Var}(\Delta \vw_0) \leq \frac{1}{N \epsilon (1 - \epsilon)} + \frac{1}{N \epsilon (1 - \epsilon)} + 2 \sqrt{\frac{1}{N \epsilon (1 - \epsilon) }}.
\]

This simplifies to:
\[
\text{Var}(\Delta \vw_0) \leq \left ( \sqrt{\frac{1}{N \epsilon (1 - \epsilon) }} + \sqrt{\frac{1}{N \epsilon (1 - \epsilon) }} \right )^2
\]

Taking the square root to get the SE:
\[
\text{SE}(\Delta \vw_0) \leq \sqrt{\frac{1}{N \epsilon (1 - \epsilon) }} + \sqrt{\frac{1}{N \epsilon (1 - \epsilon) }}.
\]

As \(N \to \infty\), both terms approach $0$, so $\text{SE}(\Delta \vw_0) \to 0$. 

\end{proof} 

Despite the restrictive assumptions, this proposition demonstrates (1) identifiability in highly simple settings to motivate future theoretical work and (2) identifies factors that contribute to the Fisher information. These factors include the number of animals pooled, $N$ (see Fig.~\ref{fig:main_default}C) and initial weights, $\vw_0$ (see Supp.\  Table~\ref{tab:w0_rmses}). We also note that the third assumption could be approximately achieved by pretraining animals to a common performance level or by pooling animals that exhibit similar initial behavior.

\newpage 
\section{Learning rules considered for simulated data} \label{scn:rules_list}

In simulated data, we model weight updates as follows: 
\begin{equation}
    \vw_{t+1} = \vw_t + \Delta \vw_t, 
\end{equation}
where \( \Delta \vw_t \) is determined by a specific ground truth learning rule. We remind readers that variables were defined in Section~\ref{scn:methods}. 

For \textbf{Markovian learning rules}, we consider the classical REINFORCE~\cite{williams1992simple} (more details in Supp.\ ~\ref{scn:reinf_details}): 
\begin{align}
    \Delta \mathbf{\vw_t} &\propto r(y_t, z_t) \nabla_{\vw} \log p(y_t \mid \vx_t, \mathbf{\vw_t}) \cr 
    &= r(y_t, z_t) \epsilon_{y_t} (1 - p_{y_t}) x_t, 
\quad \epsilon_{y_t=R} = +1, \quad \epsilon_{y_t=L} = -1, 
\end{align} 
where $p_{y_t} := p(y_t \mid \vx_t, \mathbf{\vw_t})$. 

Results for simulated data using the classical REINFORCE rule were shown in Fig.~\ref{fig:main_default}. A key property of this rule—captured well by our method—is the clear difference in weight updates following correct versus incorrect choices. To further test our approach, we considered an alternative rule that lacks this characteristic: a maximum likelihood rule (Supp.\  Fig.~\ref{fig:supp_predMax}). This can be viewed as analogous to supervised learning, where the correct label is known and the weight update aims to increase the probability of choosing the correct action: 
\begin{align}
    \Delta \mathbf{\vw_t} &\propto \nabla_{\mathbf{\vw_t}} \log p(z_t \mid \vx_t, \vw) \cr 
    &= \epsilon_{z_t} (1 - p_{z_t}) x_t,
\quad \epsilon_{z_t=R} = +1, \quad \epsilon_{z_t=L} = -1, 
\end{align} 
which doesn't depend on the action and reward. 

For \textbf{non-Markovian learning rule} in Fig.~\ref{fig:main_nonMarkovian}, we consider a modified REINFORCE with an "elibility-trace-like" factor (see Supp.\ \ref{scn:etrace_details}):   
\begin{equation}
    \Delta \mathbf{\vw_t} \propto r(y_t, z_t) \sum_s \epsilon_{y_{t-s}} (1 - p_{y_{t-s}}) x_{t-s}.  
\end{equation} 

\subsection{REINFORCE} \label{scn:reinf_details}

We present the classical REINFORCE~\cite{williams1992simple}:
\begin{align} \label{eqn:reinforce}
    \Delta \vw_t &\propto r_t(y_t, z_t) \nabla_{\vw_t} log P(y_t | x_t, \vw_t), 
\end{align}
which is derived as follows. 

We want to maximize the expected reward (expectation across the agent's stochastic behavior) by computing its gradient: 
\begin{align}
    \nabla_{\vw_t} \mathbb{E}_{P(y_t | x_t, \vw_t)} [ r(y_t, z_t)] &=  \nabla_{\vw_t} \int r(y_t, z_t) P(y_t | x_t, \vw_t) dy_t \cr 
    &\overset{(a)}{=} \int r(y_t, z_t) \nabla_{\vw_t} P(y_t | x_t, \vw_t) dy_t \cr 
    &\overset{(b)}{=} \int r(y_t, z_t) P(y_t | x_t, \vw_t) \nabla_{\vw_t} log P(y_t | x_t, \vw_t) dy_t \cr 
    &= \mathbb{E}_{P(y_t | x_t, \vw_t)} [ r(y_t, z_t) \nabla_{\vw_t} log P(y_t | x_t, \vw_t)], 
\end{align}
where $(a)$ is because $r(y_t,z_t)$ is independent of $\vw_t$ once $y_t$ is given, and $(b)$ is from the log-derivative trick. We can evaluate the right-hand-side with a Monte Carlo integral by assuming the animal’s choice $y_t$ is sampled according to its policy $P(y_t | x_t, \vw_t)$, which results in Eq.~\ref{eqn:reinforce}. 

\subsection{REINFORCE with "eligibility trace"} \label{scn:etrace_details}

Suppose we change the task setting and the reward $r_t$ now depends up to past $S$ data points, i.e. $r_t=r_t(y_{t-S:t}, z_{t-S:t})$, then:
\begin{align}
    &\nabla_{\vw_t} \mathbb{E}_{P(y_{t-S:t} | x_{t-S:t}, \vw_{t-S:t})} [ r(y_{t-S:t}, z_{t-S:t})] \cr &=  \nabla_{\vw_t} \int r(y_{t-S:t}, z_{t-S:t}) P(y_{t-S:t} | x_{t-S:t}, \vw_{t-S:t}) dy_{t-S:t} \cr 
    &\overset{(a)}{=} \int r(y_{t-S:t}, z_{t-S:t}) \nabla_{\vw_t} P(y_{t-S:t} | x_{t-S:t}, \vw_{t-S:t}) dy_{t-S:t} \cr 
    &\overset{(b)}{=} \int r(y_{t-S:t}, z_{t-S:t}) P(y_{t-S:t} | x_{t-S:t}, \vw_{t-S:t}) \nabla_{\vw_t} log P(y_{t-S:t} | x_{t-S:t}, \vw_{t-S:t}) dy_{t-S:t} \cr 
    &= \mathbb{E}_{P(y_{t-S:t} | x_{t-S:t}, \vw_{t-S:t})} [ r(y_{t-S:t}, z_{t-S:t}) \nabla_{\vw_t} log P(y_{t-S:t} | x_{t-S:t}, \vw_{t-S:t})], \cr 
    &\overset{(c)}{\approx} \mathbb{E}_{P(y_{t-S:t} | x_{t-S:t}, \vw_{t-S:t})} [ r(y_{t-S:t}, z_{t-S:t}) \sum_{s=0}^S \nabla_{\vw_{t-s}} log P(y_{t-s} | x_{t-s}, \vw_{t-s})], 
\end{align}
where $(a)$ is because $r(y_{t-S:t},z_{t-S:t})$ is independent of $\vw$ once $y_{t-S:t}$ is given, and $(b)$ is from the log-derivative trick; $(c)$ we assume $\vw_{t-s}\approx \vw_t$ and also $P(y_{t-S:t} | x_{t-S:t}, \vw_{t-S:t})=\Pi_{s=0}^S P(y_{t-s} | x_{t-s}, \vw_{t-s})$ from the task setup. Again, we apply Monte-Carlo approximation:  
\begin{align} \label{eqn:n_step}
    \Delta \vw_t &\propto r_t \sum_{s=0}^S \nabla_{\vw_{t-s}} log P(y_{t-s} | x_{t-s}, \vw_{t-s}). 
\end{align} 

For Fig.~\ref{fig:main_nonMarkovian}, we used $S=10$ and defined $r_t=r_t(y_{t-S:t}, z_{t-S:t})$ as $1$ if at least half of the recent $S$ trials are correct and $0$ otherwise; we found similar results for other choices too. We refer to it as 'eligibility-trace-like', although it may not arise from eligibility traces, since the reward is combined with an input/output-dependent factor that accumulates over multiple time steps. 

\newpage 
\section{Additional details on model and data} \label{scn:sim_details}

\textbf{Simulated data.} We simulate a pool of animals, each following the same underlying learning rule. In Supp.\  Fig.~\ref{fig:supp_mixed_rates}, we also explore populations with mixed weight update functions, where pooling tends to recover an in-between weight update function. Even when animals follow the same rule, their learning trajectories differ due to the stochasticity of received inputs and behavior.

For each simulated animal, the learning process unfolds over trials $t = 0, 1, \dots, T$ as follows. Each stimulus $s_t$ is drawn uniformly from discrete values in $[-2, 2]$ with 0.25 increments, excluding 0 (except for Fig.~\ref{fig:main_default}, which includes 0); varying the increment size did not affect our conclusions. The correct label or answer is defined as $z_t = \indicator[s_t > 0]$. Decision weights are updated according to $\vw_{t+1} = \vw_t + \Delta \vw_t$, where $\Delta \vw_t$ is computed from the ground-truth learning rule being tested (see Supp.\ ~\ref{scn:rules_list}). The agent's binary choice $y_t \in \{0, 1\}$ is sampled from a Bernoulli GLM parameterized by $\vw_t$, as described in Section~\ref{scn:methods}. A reward is given if the choice matches the correct answer: $r_t = \indicator[y_t = z_t]$. 

For the total number of trials $T$, we use $T = 500$ by default (to speed up simulations), but also demonstrate that our conclusions hold for longer $T$ ($T = 8000$) in Supp.\  Fig.~\ref{fig:supp_T8000}. We set the learning rate $\alpha$ such that the animal's performance reaches a saturation regime --- typically when the stimulus weight approaches a value of $3$ for this task. 
The GLM includes a stimulus weight that is updated over trials, and a fixed bias weight sampled randomly at the start of each simulation from $\{-1, 0, 1\}$. We simulate a pool of 10 to 1000 animals to observe the influence of data availability on reconstruction accuracy in Fig.~\ref{fig:main_default}C. We evaluate model performance using 5-fold cross-validation, where each fold holds out a subset of animals entirely. This ensures generalization across individuals, consistent with the underlying assumption of pooling --- that animals share commonalities which can be learned from a subset and applied to held-out animals. 

\textbf{IBL data.} We use the publicly available International Brain Laboratory (IBL) dataset~\cite{international2020standardized}, and follow the same preprocessing pipeline as~\cite{ashwood2020inferring}, with details available in their accompanying code. For dataset details and learning curves, refer to Supp.\  C and Fig.\ S1 in~\cite{ashwood2020inferring}. In other words, we used the same preprocessed dataset as~\cite{ashwood2020inferring} ($N=12$), for benchmarking purposes. For evaluation, we adopt two cross-validation strategies, and both led to similar trends (Tables~\ref{tab:heldout-LL-results} and~\ref{tab:future_LL}). First, under the pooling assumption that a shared learning model generalizes across animals, we use $K$-fold cross-validation with animal-held-out splits. We fit the model using the first 10000 trials per animal to capture the full learning trajectory, including the saturation phase. For each fold, we report the average validation log-likelihood across four random seeds and conduct paired $t$-tests across matched animals to obtain the $p$-values reported in Table~\ref{tab:heldout-LL-results}. The paired design accounts for variability in individual animals’ performance: animals who learn well are easier to predict, while animals who have not learned the task are harder to model (see Fig. S1 in~\cite{ashwood2020inferring}). Second, to assess temporal generalization, we adopt a future-data holdout strategy~\cite{hyndman2018forecasting}. We train up to the point where the learning curve inflects (around trial 5500), ensuring that the evaluation data reflect ongoing learning rather than the saturation regime. This also avoids cutting off training too early, where little learning signals are present. We then evaluate performance on the next 500 trials beyond that point, although similar trends were observed when using 200 or 1000 trials instead. As above, we compute paired $t$-tests on animal-matched log-likelihoods to assess significance. 

\textbf{Learning model, initial GLM weights, and hyperparameters.} 
Our learning rule model is depicted in Fig.~\ref{fig:main_schematics}C. The network takes as input the current trial stimulus $s_t$, choice $y_t$, reward $r_t$, and the current weight $\vw_t$. For the IBL data, we additionally include binary inputs indicating daybreak and stimulus side; the trends in Table~\ref{tab:heldout-LL-results} remain unchanged when these inputs are omitted. The network then outputs inferred weight update. For choice and reward, we input the animal's actual choice and reward rather than those generated by the model, since these are the signals the animal would have access to when updating its weights; this approach is also consistent with previous studies. The DNN architecture consists of two hidden layers with 32 units each, while the RNN uses a single recurrent layer with 32 hidden units. All architectural components and training settings are treated as hyperparameters and tuned via cross-validation by animal held-out. All neural network weights and biases were treated as trainable parameters. 

To estimate the initial decision weights $\vw_0$, we treat them as trainable parameters but initialize them using a psychometric curve fit to the first 100 trials. Specifically, we compute the empirical probability of choosing $y = 1$ as a function of stimulus value, which defines a psychometric curve. Since we model the choice probability as $\sigma(\vw^\top x)$, where $\sigma$ is the logistic sigmoid, we apply the inverse-sigmoid (logit) transformation to the empirical probabilities, yielding a linear target $b = \text{logit}(p(y=1|x))$. We then fit $\vw_0$ via a least-squares regression to the equation $\vw^\top x = b$ using the stimulus values. We investigate the sensitivity of performance to incorrect $\vw_0$ in Table~\ref{tab:w0_sensitivity}. As mentioned, to enable faster simulations and repeated robustness tests, we used shorter trial sequences. Because these provide fewer early samples, we initialized with a known initial weight. That said, applying our method to unknown initial weight still led to accurate recovery (Supp.\ Fig.~\ref{fig:supp_T8000}). For IBL data, however, we instead initialize $\vw_0 = \vec 0$, following~\cite{ashwood2020inferring}, since animals typically start at chance-level performance; we also verified that training $\vw_0$ did not significantly impact our results on IBL data. 

We tune hyperparameters, including the number of hidden units, number of layers, learning rate, and number of training epochs. We used a batch size of 1 for simulated data. For real IBL data, we trained on all animals jointly; using a batch size of 1 instead did not affect our conclusions. Cross-validation ensures generalization and helps mitigate overfitting, in addition to the fact that neural networks in feature learning regime inherently favor smooth function approximators via implicit regularization (see Discussion). These implicit biases are themselves controlled by architectural choices and training regimes (e.g., depth, width, and training duration)~\cite{wilson2025deep,geiger2020disentangling}, which justifies the need for hyperparameter tuning. While we do not apply any explicit regularization in this work, exploring the interplay between explicit and implicit regularization on learning rule identifiability remains an interesting direction for future research. Validation-driven model selection is crucial to avoid recovering an incorrect learning rule that merely overfits the training data --- achieving high training log-likelihood but poor generalization to held-out data.  

\textbf{Computational details.}
Our code is available \url{https://github.com/Helena-Yuhan-Liu/InferLearningANNGLM}. We implemented our models in PyTorch v1.13.1~\cite{paszke2019pytorch}. Simulations were run on a machine equipped with a 12th Gen Intel(R) Core(TM) i7-12700H processor (14 cores, 20 threads) with a maximum clock speed of 2.30GHz. Training averages to around 2 hours per run. We used the Adam optimizer with a default learning rate of $1e-3$, and binary cross-entropy loss, which corresponds to the negative log-likelihood under a Bernoulli model. For statistical analysis and significance testing, we used tools from the SciPy package~\cite{virtanen2020scipy}.

\newpage 
\section{Additional results}

\begin{figure}[h!]
    \centering
    \includegraphics[width=0.98\textwidth]{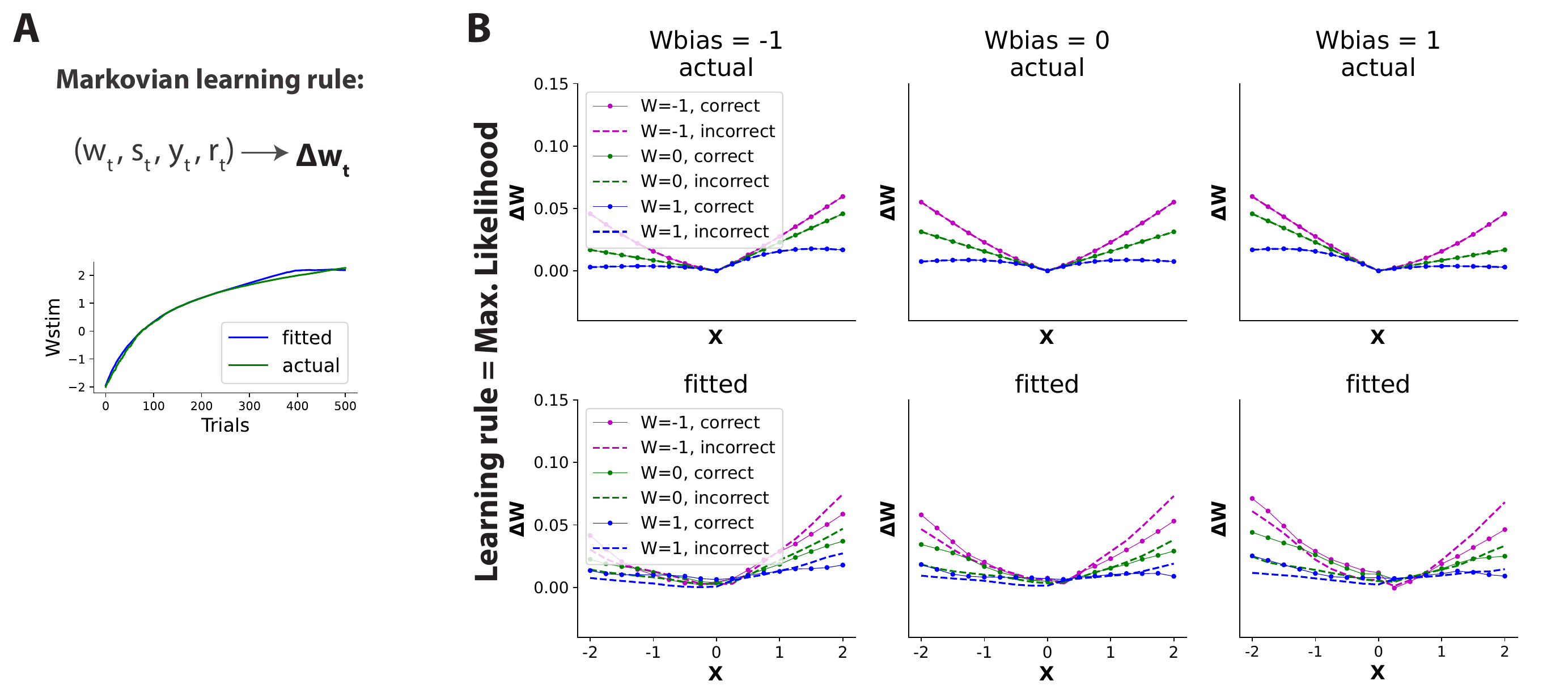}
    \caption{ Similar to Fig.~\ref{fig:main_default} but repeated for Max. Likelihood as the ground truth in simulated data. Again, we see that the reconstruction successfully captures key characteristics: the slowing of learning at higher weights, the increase with stimulus amplitude, and the lack of clear separation between correct versus incorrect decisions. 
    } \label{fig:supp_predMax}
\end{figure}

\begin{figure}[h!]
    \centering
    \includegraphics[width=0.98\textwidth]{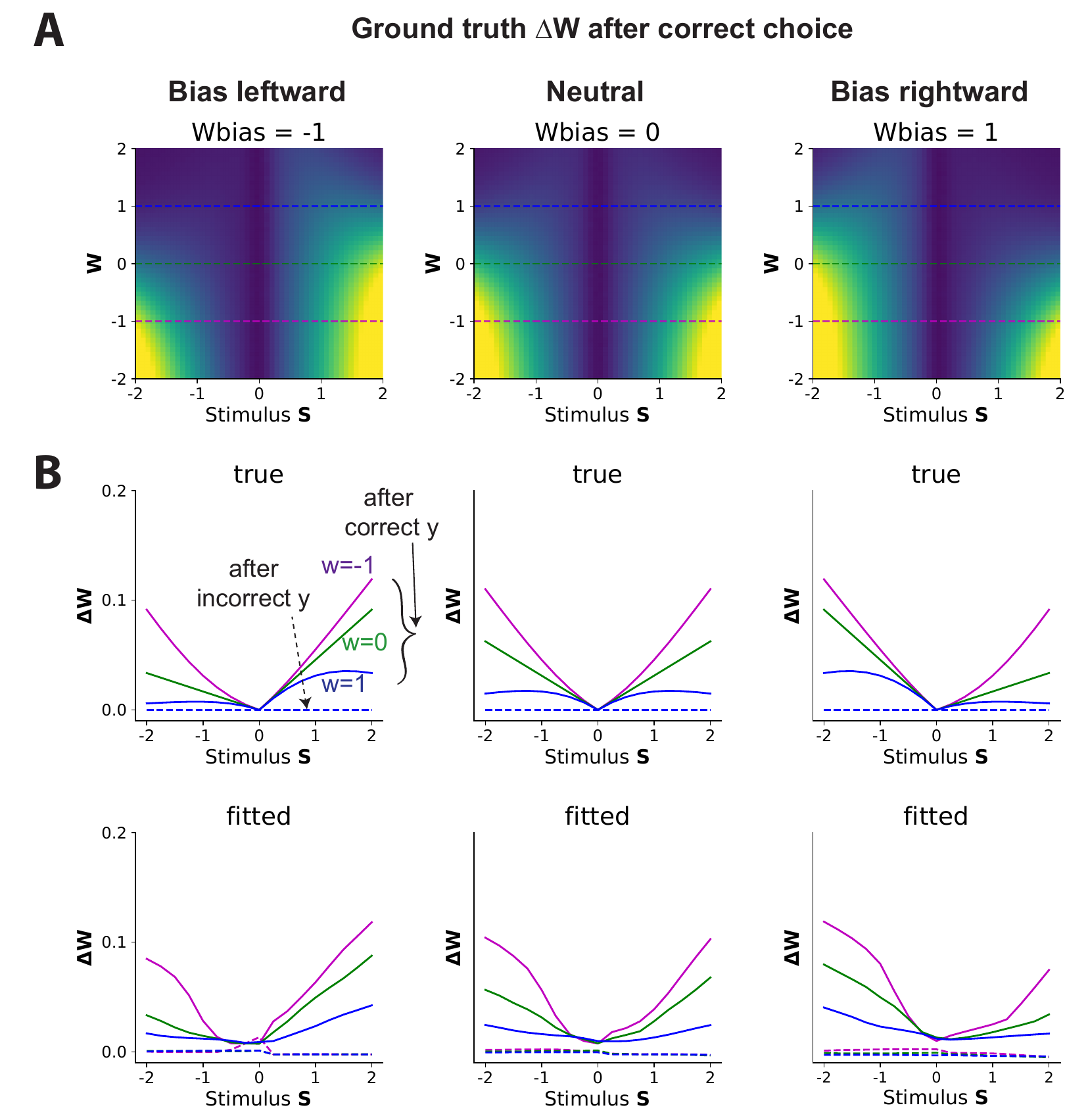}
    \caption{ Similar to Fig.~\ref{fig:main_default}, but repeated while holding at other bias weight values in simulated data. The trends in Fig.~\ref{fig:main_default} hold, and our reconstruction also captures the side asymmetry when the bias weight is nonzero. 
    } \label{fig:supp_default}
\end{figure}

\begin{figure}[h!]
    \centering
    \includegraphics[width=0.9\textwidth]{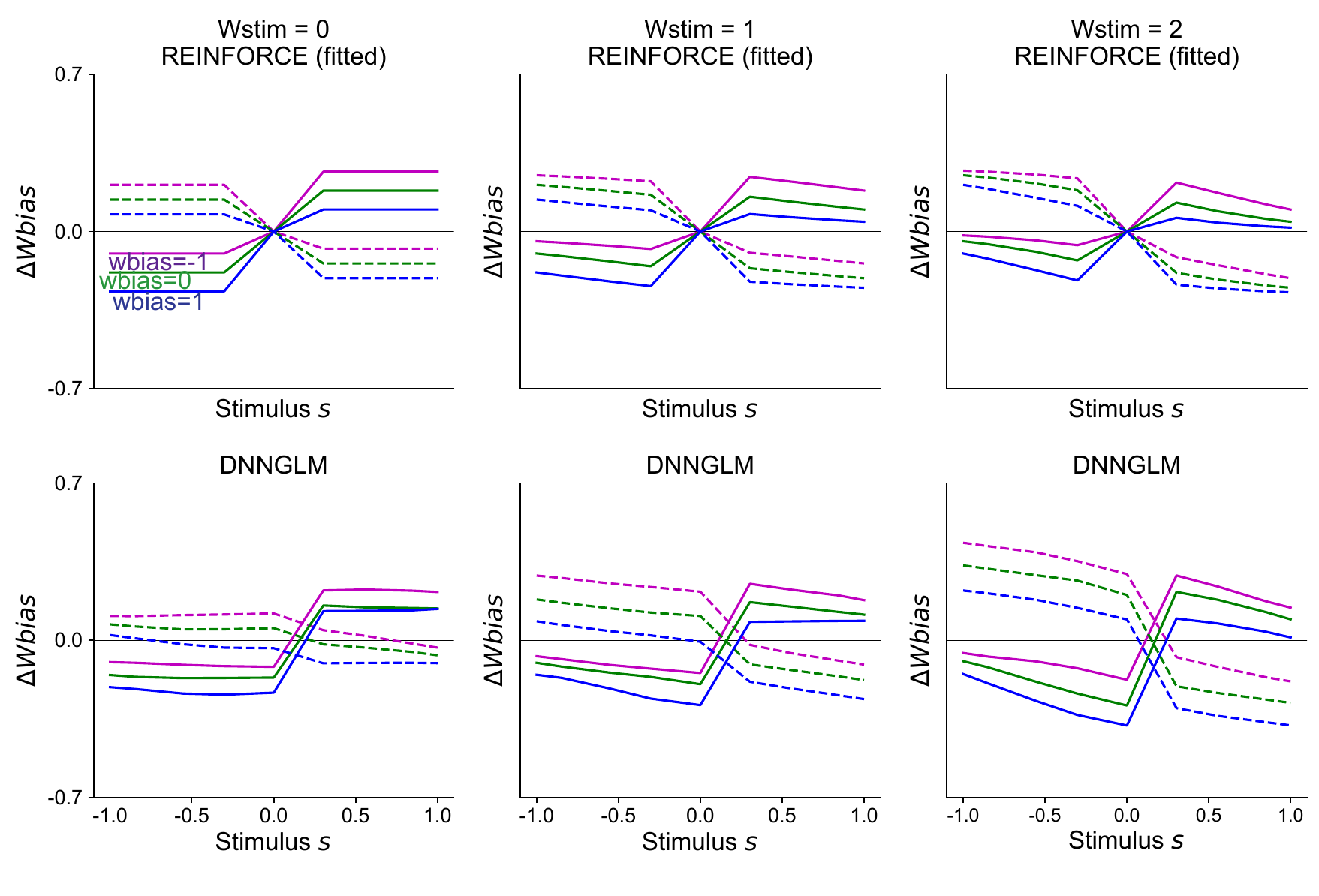}
    \caption{ \textbf{Bias weight update function closely resembles REINFORCE.} Similar to Fig.~\ref{fig:main_ibl2}, which visualizes the stimulus weight update function, we plot here the inferred update function for the bias weight. We plot REINFORCE (fitted on the DNNGLM) in the top row and DNNGLM on the bottom row. The fitted neural network model captures key qualitative features of the REINFORCE update rule for the bias weight. }
 \label{fig:supp_IBL_db}
\end{figure}

\begin{figure}[h!]
    \centering
    \includegraphics[width=0.98\textwidth]{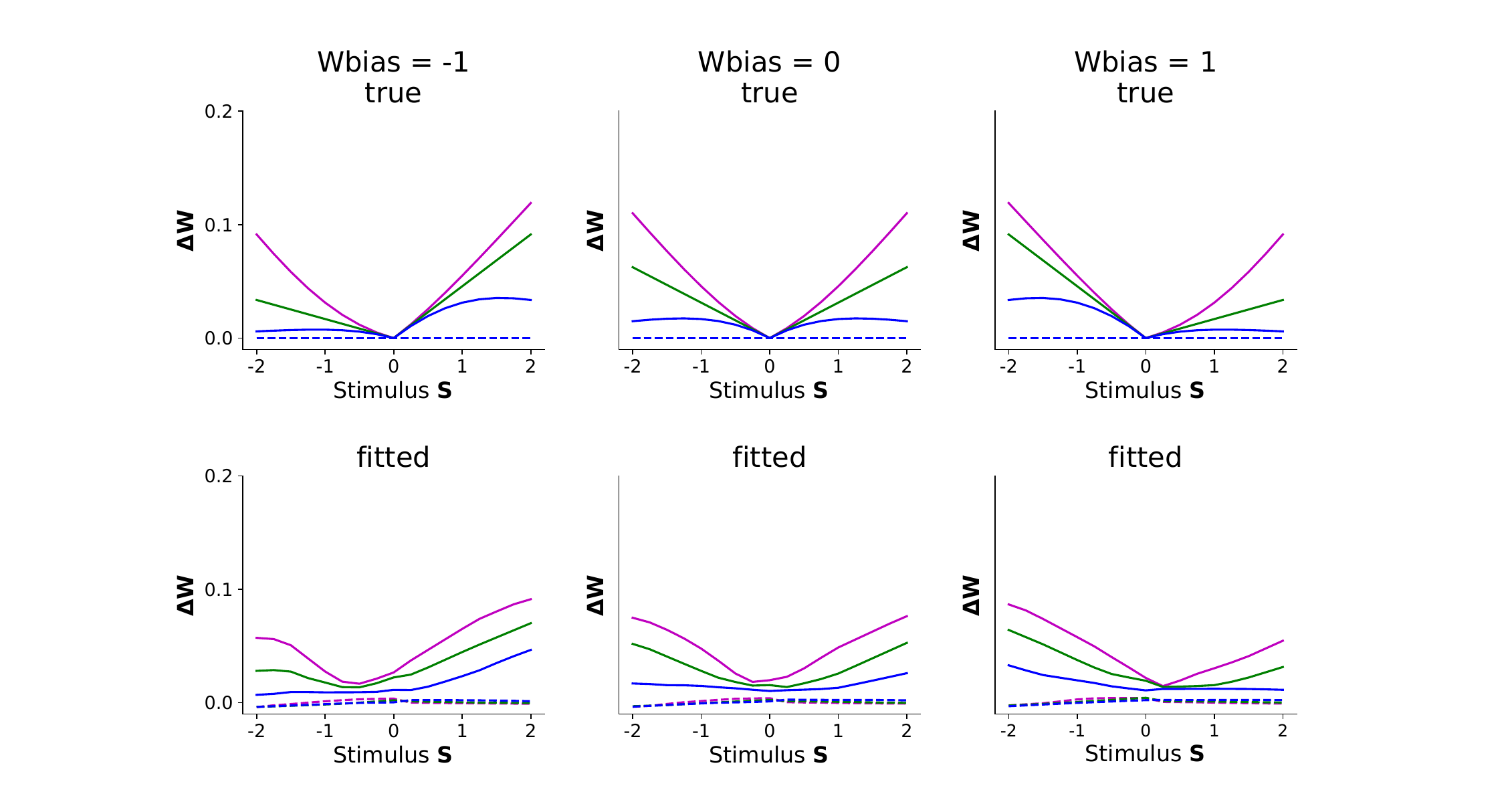}
    \caption{ \textbf{Different initial stimulus weight \( \vw_0 \).} 
\textbf{(A)} In our main results for simulated data, we initialized with stimulus weight \( \vw_0 = -2 \) so that the animal starts with poor performance and learns over time, thereby covering a wide range of stimulus weights during training. Here, we instead randomly initialize \( \vw_0 \sim \mathcal{U}[-2, 2] \) (so different animals begin with different initial weights) and observe similar overall trends in the learned weight update function. Plotting convention follows that of Fig.~\ref{fig:main_default}. 
} \label{fig:supp_unifW0}
\end{figure}

\begin{figure}[h!]
    \centering
    \includegraphics[width=0.98\textwidth]{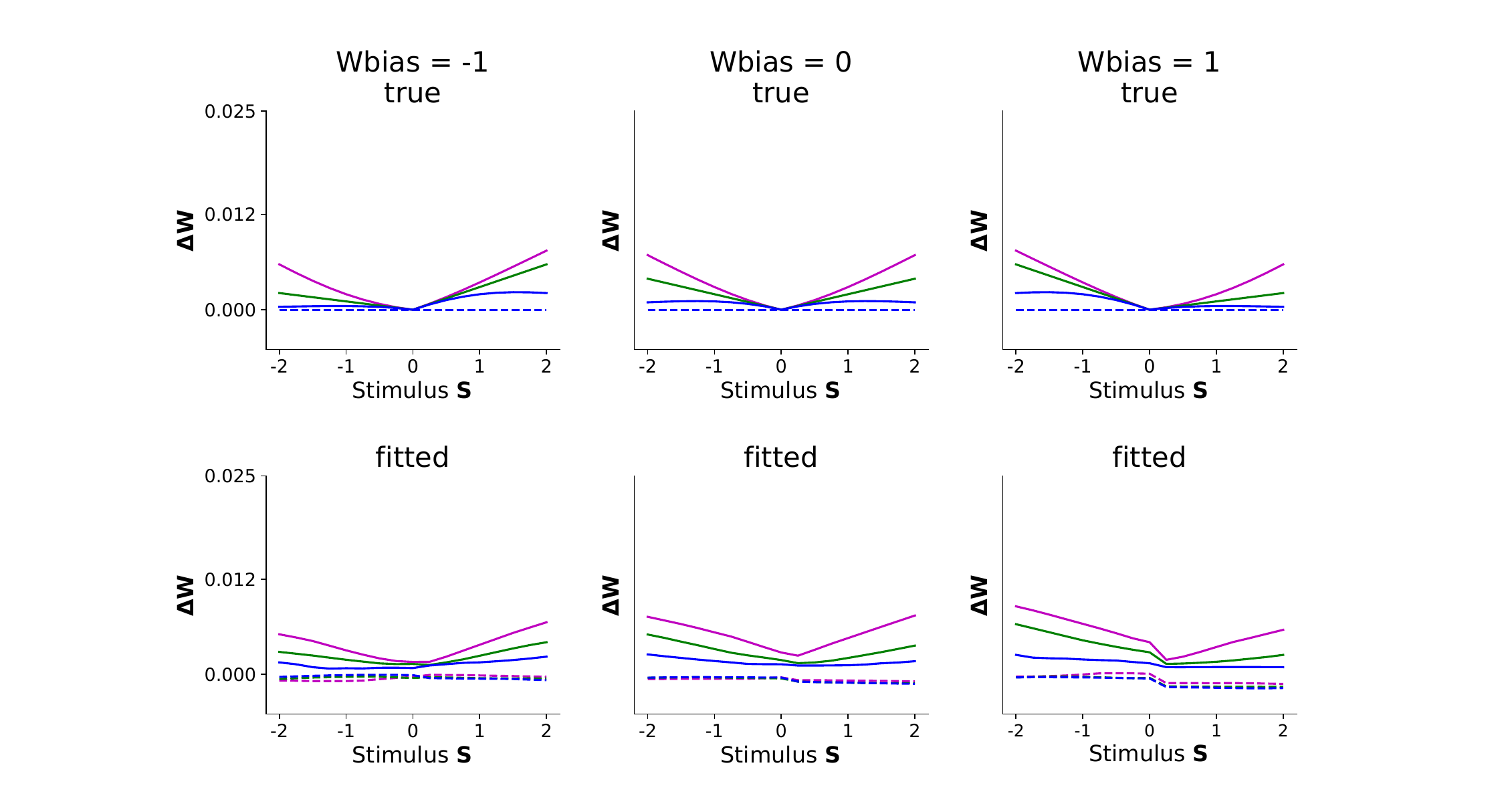}
    \caption{ Trends observed in Fig.~\ref{fig:main_default} hold for longer trial sequences ($T=8000$). Plotting conventions are identical to those in Fig.~\ref{fig:main_default} for simulated data. 
} \label{fig:supp_T8000}
\end{figure}

\begin{figure}[h!]
    \centering
    \includegraphics[width=0.98\textwidth]{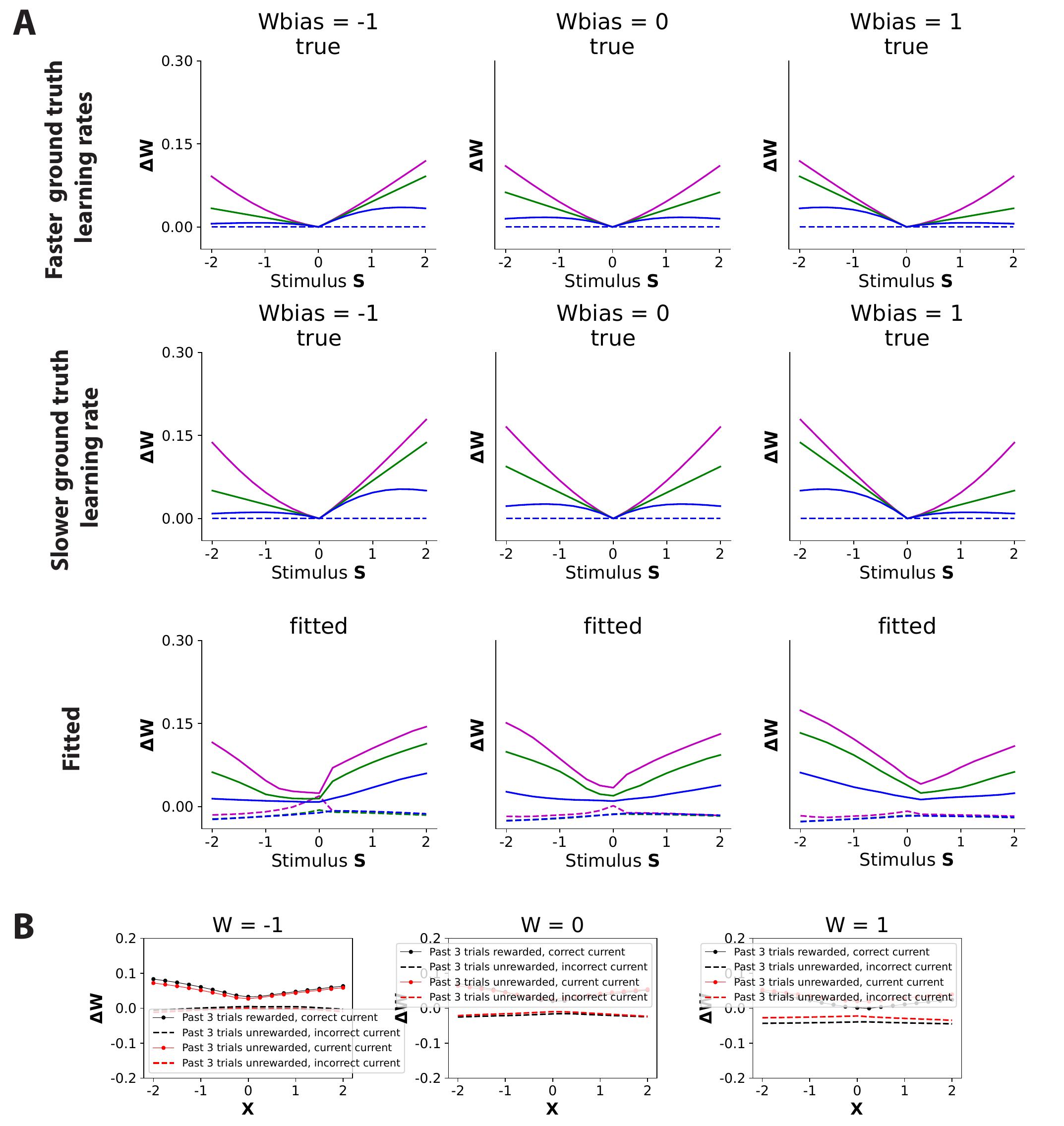}
    \caption{ \textbf{Pooling with mixed weight update functions.} 
\textbf{(A)} Following the plotting convention in Fig.~\ref{fig:main_default}, we simulate a mixed population where half the animals learn with REINFORCE using a higher learning rate (top row) and the other half with a lower learning rate (middle row). The fitted function (bottom row) recovers an intermediate update rule that lies between the two subpopulations. 
\textbf{(B)} To test whether the reward-history dependence observed in Fig.~\ref{fig:main_ibl2} is merely a result of faster-learning animals in the pool contributing more recently rewarded trials, we train RNNGLM on a population with mixed learning rates. Reward-conditioned weight update plots --- where past trial rewarded (black) and unrewarded (red) --- do not consistently show the reward history dependence in this setting. This suggests that the history dependence learned by RNNGLM in IBL data is not merely due to faster learners receiving more recent rewards; this makes sense as in both this plot and Fig.~\ref{fig:main_ibl2}, the learning stage is controlled by fixing the stimulus weight $\vw$. 
} \label{fig:supp_mixed_rates} 
\end{figure}

\begin{figure}[h!]
    \centering
    \includegraphics[width=0.9\textwidth]{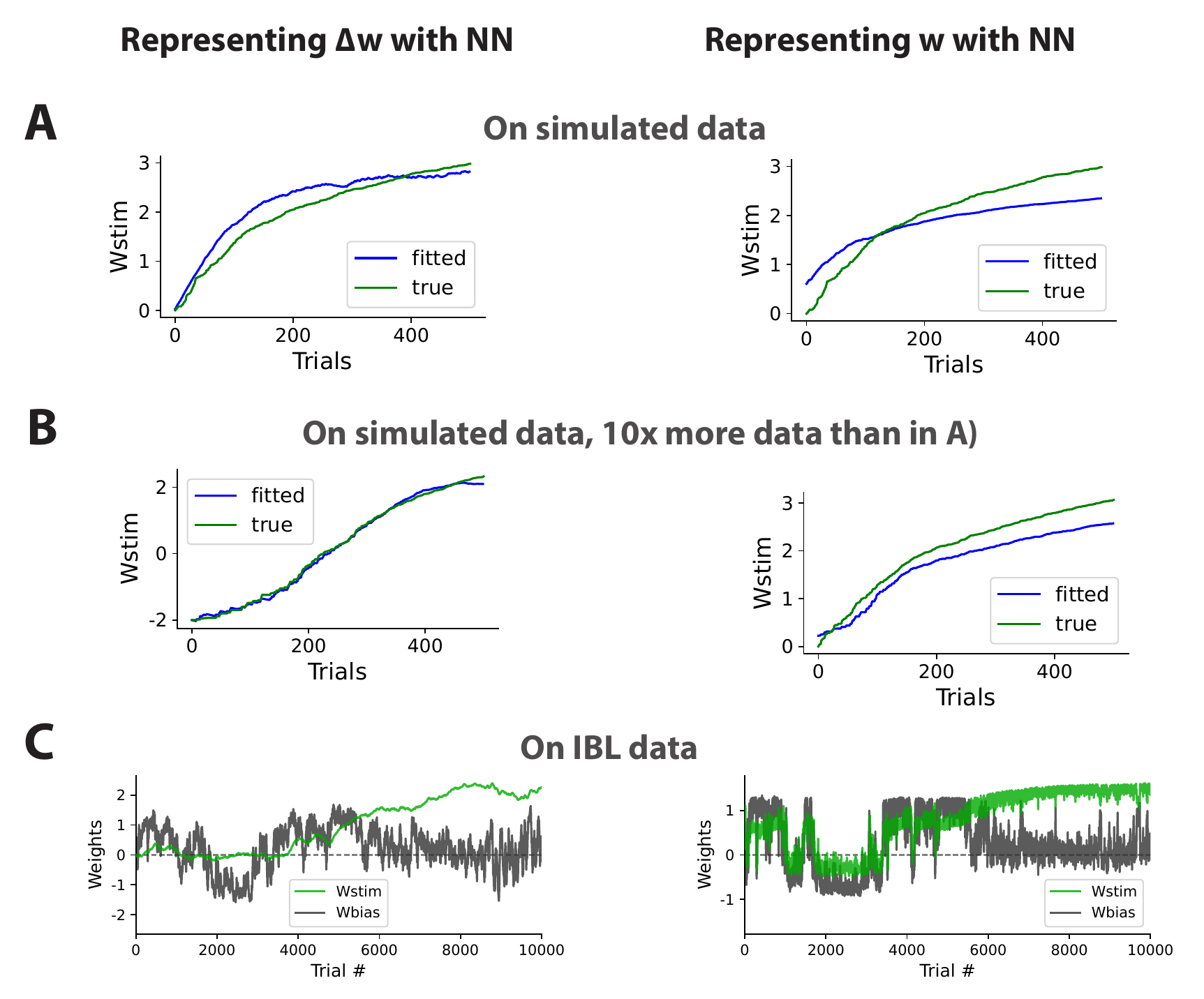}
    \caption{\textbf{Parameterizing $\Delta \vw$ directly (as opposed to $\vw$) improves weight trajectory recovery.} 
\textbf{(A)} On simulated data, using a neural network to directly represent the weight update $\Delta \vw$ yields more accurate recovery of the weight trajectory compared to directly parameterizing $\vw$ using an RNN. The RNN-based $\vw$ trajectory is overly smooth, likely due to implicit regularization that biases the model toward simpler solutions in the weight trajectory rather than the weight update function; in contrast, regularizing the weight changes (via $\Delta \vw$) provides a more appropriate inductive bias (see Discussion). 
\textbf{(B)} As expected, the effect of implicit regularization is more pronounced in underconstrained settings (e.g., low-data regimes), and becomes less critical as more data or constraints are introduced; since with sufficient data, both models should approximate the learning trajectory well as per the universal approximation results. 
\textbf{(C)} On IBL data, this distinction is also evident: when training on just one example animal, parameterizing $\Delta \vw$ leads to more realistic weight trajectories than using an RNN to represent $\vw$ directly. 
} \label{fig:supp_RNNW}
\end{figure}

\begin{figure}[h!]
    \centering
    \includegraphics[width=0.98\textwidth]{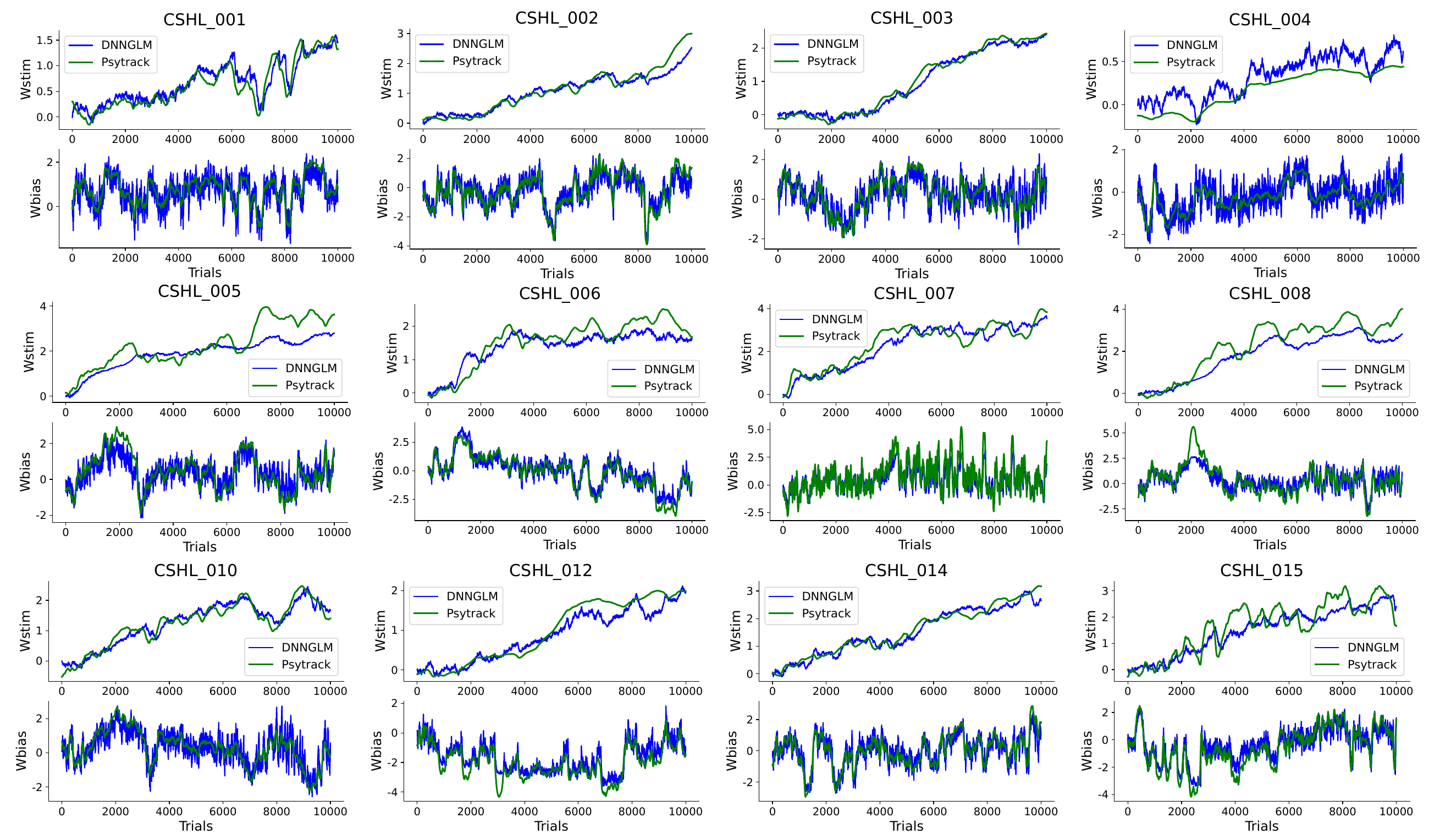}
    \caption{ Extended version of Fig.~\ref{fig:main_ibl}A, showing both stimulus and bias weight trajectories across more animals. The decision weights recovered by DNNGLM closely resemble those inferred by PsyTrack. 
} \label{fig:supp_psyDNN}
\end{figure}

\begin{figure}[h!]
    \centering
    \includegraphics[width=0.98\textwidth]{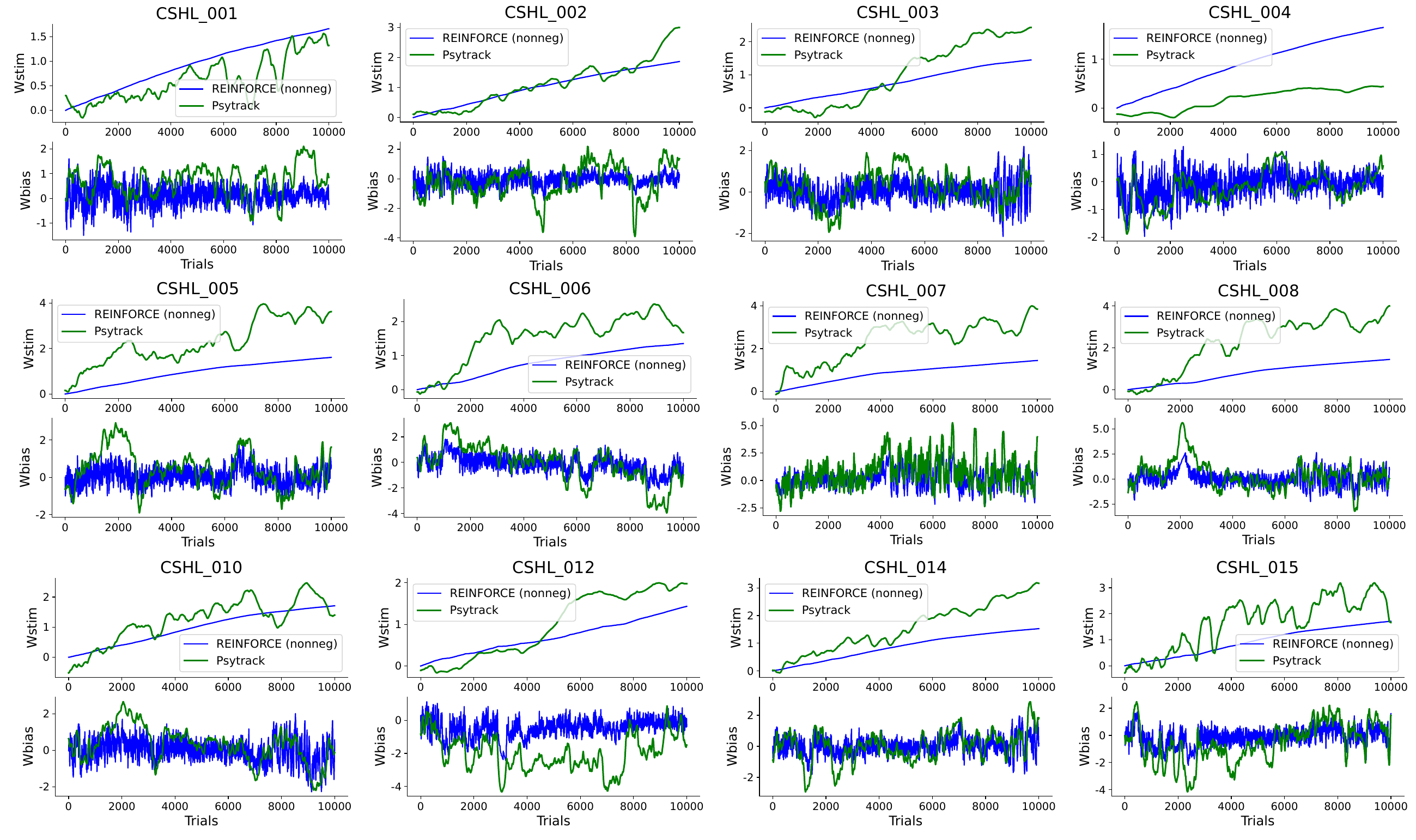}
    \caption{ Similar to Supp.\  Fig.~\ref{fig:supp_psyDNN}, but showing the decision weights recovered by REINFORCE (with the nonnegative baseline constraint), which was previously shown to yield worse log-likelihoods (Tables~\ref{tab:heldout-LL-results} and~\ref{tab:future_LL}). We see here that it leads to greater discrepancies from the weights inferred by PsyTrack.
} \label{fig:supp_psyReinf}
\end{figure}

Fig.~\ref{fig:supp_default} repeats Fig.~\ref{fig:main_default} for additional bias weight values. Fig.~\ref{fig:supp_IBL_db} shows the fitted bias weight updates for the IBL dataset. Table~\ref{tab:future_LL} echoes the trends in Table~\ref{tab:heldout-LL-results} with future data heldout instead of animal-level heldout. Additionally, we also demonstrate our main findings are robust across a variety of conditions: (i) a learning rule without the decision outcome asymmetry (Supp.\  Fig.~\ref{fig:supp_predMax}); (ii) recovery of an in-between learning rule when mixing update functions across individuals (Supp.\  Fig.~\ref{fig:supp_mixed_rates}); (iii) different initial stimulus weights (Supp.\  Fig.~\ref{fig:supp_unifW0}); (iv) longer trial lengths (Supp.\  Fig.~\ref{fig:supp_T8000}). We also show sensitivity to the initial weight estimate (Supp.\  Table~\ref{tab:w0_rmses}) and degradation of recovery under added noise and incorrect initial weight estimates (Supp.\  Table \ref{tab:rmse_noise} and~\ref{tab:w0_sensitivity}). Comparisons for parameterizing the weights $\vw$ versus the updates $\Delta \vw$ are provided in Fig.~\ref{fig:supp_RNNW}. Finally, we show the uncovered weight trajectories in Fig.~\ref{fig:supp_psyDNN} and~\ref{fig:supp_psyReinf}. 

During initial rebuttal experiments, we observed slightly higher performance for the Transformer variant compared to the RNN under limited training epochs. In the final analysis with full training epochs, this trend no longer holds — performance across architectures is comparable ($p=7.44e-1$). We note that RNN-Transformer comparison is not central to our main contribution. We hypothesize that the comparable performance may be due to the small sample size (12 animals, as in~\cite{ashwood2020inferring}). We anticipate that the scalability advantages of Transformer and state-space models (e.g. Mamba) will become increasingly important as we extend to larger datasets. 

\newpage 

\newpage 

\newpage 

\begin{table}[h]
\centering
\begin{tabular}{lcc}
\toprule
\textbf{Comparison} & \textbf{Heldout LL} & \textbf{$p$-val} \\
\midrule
REINFORCE vs. \textbf{DNNGLM}                     & $-2924.4$ vs. $\mathbf{-2905.8}$ & $1.41\text{e}{-2}$ \\
DNNGLM vs. \textbf{RNNGLM}                        & $-2905.8$ vs. $\mathbf{-2881.2}$ & $2.24\text{e}{-3}$ \\
\textbf{REINFORCE} vs. nonnegative base           & $\mathbf{-2924.4}$ vs. $-3272.4$ & $1.08\text{e}{-9}$ \\ 
\midrule
REINFORCE (history) vs. \textbf{RNNGLM}           & $-2914.8$ vs. $\mathbf{-2881.2}$ & $5.61\text{e}{-4}$ \\
DNNGLM (history) vs. \textbf{RNNGLM}              & $-2890.2$ vs. $\mathbf{-2881.2}$ & $3.52\text{e}{-2}$ \\
tinyRNN vs. \textbf{RNNGLM}                       & $-2893.8$ vs. $\mathbf{-2881.2}$ & $3.38\text{e}{-2}$ \\
\bottomrule
\end{tabular}
\caption{
Future-data prediction log-likelihoods (higher is better) for different learning models, along with $p$-values from paired $t$-tests matched across mice. These results mirror the trends observed in Table~\ref{tab:heldout-LL-results}. 
}
\label{tab:future_LL}
\end{table}

\begin{table}[h!]
\centering
\begin{tabular}{c c}
\hline
Initial stimulus weight $\vw_0$ & RMSE of reconstructed $\Delta \vw$ \\
\hline
$-2$ & 0.0130 \\
$0$ & 0.0244 \\
$2$ & 0.0287 \\
\hline
\end{tabular}
\caption{\textbf{Impact of initial stimulus weight $\vw_0$ on reconstruction accuracy.} 
We measure reconstruction RMSE of the inferred $\Delta \vw$ when varying the ground truth initial weight $\vw_0$ in simulated data, keeping all other factors fixed. 
Initializing with $\vw_0 = -2$ yields the lowest error, while $\vw_0 = 2$ leads to saturation and poor identifiability. This pattern is consistent with Fisher Information analysis (Supp.\ \ref{scn:theory}): having more datapoints around weight values near $\vw = 0$ correspond to choice probabilities near 0.5, which maximize Fisher Information and enhance identifiability of learning rules. Starting from $\vw_0 = -2$ ensures the learning trajectory passes through this region, enabling more accurate recovery of the update function. This trend holds even in light of the fact that the value of $\Delta \vw$ would be smaller for higher $\vw$ in REINFORCE. This analysis suggests that having animals initially perform poorly on the task, even below chance level, could potentially provide more information for learning rule identification.}
\label{tab:w0_rmses} 
\end{table} 

\begin{table}[h!]
\centering
\begin{tabular}{l c}
\toprule
\textbf{Noise Level} & \textbf{RMSE (Reconstructed $\Delta \vw$ vs. Ground Truth)} \\
\midrule 
$\sigma = \frac{1}{4} \alpha$ & 0.0216 \\
$\sigma = \frac{1}{2} \alpha$ & 0.0349 \\
$\sigma = 1 \alpha$           & 0.0652 \\
\bottomrule
\end{tabular}
\caption{Reconstruction accuracy degrades with increasing white Gaussian noise in simulated data. We add zero-mean white Gaussian noise to the ground truth REINFORCE learning rule for simulated data, with the noise standard deviation $\sigma$ expressed as a fraction of the learning rate $\alpha$. In other words, the ground truth weight changes are now driven by two components: a learning component and a noise component, as in~\cite{ashwood2020inferring}, whereas our fitted model remains deterministic. We report RMSE between the reconstructed and true $\Delta \vw$. As expected, higher noise leads to worse recovery accuracy.}
\label{tab:rmse_noise}
\end{table} 

\begin{table}[h]
\centering
\begin{tabular}{lc}
\toprule
\textbf{$\vw_0$ correctness} & \textbf{RMSE} \\
\midrule
Default  & 0.0130 \\
+0.5 perturb        & 0.0270 \\
+1.0 perturb        & 0.0453 \\
+2.0 perturb        & 0.1519 \\
\bottomrule
\end{tabular}
\caption{Effect of initial weight $\vw_0$ correctness on reconstruction accuracy in simulated data. We test the following scenarios: default, and $\vw_0$ offset by +0.5, +1.0, +2.0 after the estimation. Deviations from the true $\vw_0$ can substantially degrade the accuracy of learning rule recovery, underscoring the importance of accurately estimating $\vw_0$.}
\label{tab:w0_sensitivity}
\end{table}

\begin{table}[h]
\centering
\begin{tabular}{lccc}
\toprule
Additional Input Dim & Default Sample Size & 3$\times$ Samples \\
\midrule
1   & 1.00 & 0.59 \\
3   & 1.09 & 0.62 \\
10  & 1.35 &  0.7 \\
30  & 1.58 &  1.20 \\
100 &  X & 1.67 \\
\bottomrule
\end{tabular}
\caption{\textbf{Recovery performance (relative RMSE) of the inferred learning rule across increasing input dimensionality.} 
Each row corresponds to an input dimensionality setting; additional inputs are distractor streams added to the default task in Fig.~\ref{fig:main_default}. 
An “X” marks failure to capture the qualitative features in Fig.~\ref{fig:main_default}; features remain captured at higher-D task settings, and reconstruction accuracy improves with more data. Relative RMSE is normalized to the baseline ($1$ additional input, default sample size).
}
\label{tab:input-dim}
\end{table} 

\begin{table}[h]
\centering
\begin{tabular}{l c}
\toprule
Condition & Relative RMSE \\
\midrule
Decision model known & 1.00 \\
Model mismatch       & 1.62 \\
\bottomrule
\end{tabular}
\caption{ \textbf{Performance degradation with decision model mismatch.} We examine a multiplicative decision structure as the ground-truth model. 
The first row corresponds to the case where the inference framework knows the true multiplicative decision structure, 
$\text{logit}(P(\text{right})) = w_L \cdot c \cdot x_L + w_R \cdot (1 - c) \cdot x_R$, 
for rightward cumulative cue and context variable $c$ between 0 and 1, and leftward cumulative cue between 0 and -1. 
The second row represents model mismatch, where the inference framework still assumes a GLM model, 
$\text{logit}(P(\text{right})) = \hat{w}_L \cdot x_L + \hat{w}_R \cdot x_R + \hat{w}_c \cdot c$.}
\label{tab:multiplicative-decision}
\end{table}

\begin{figure}[h!]
    \centering
    \includegraphics[width=0.9\textwidth]{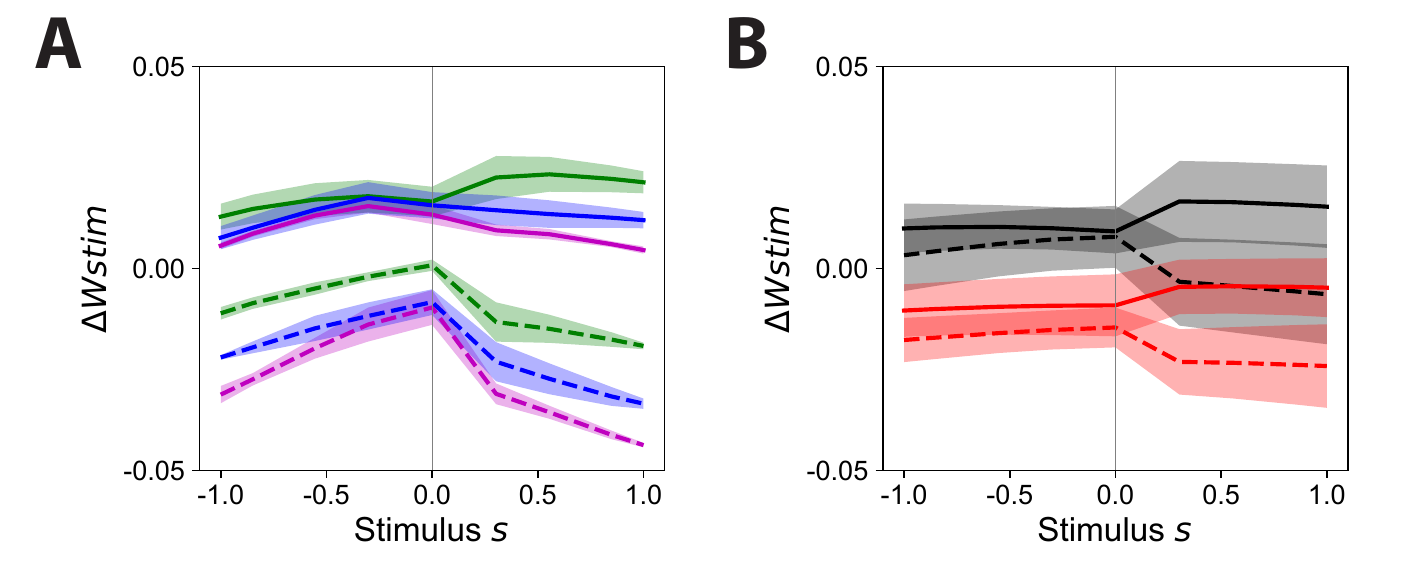}
    \caption{\textbf{Robustness of inferred learning rules across random seeds.} 
We ran all inference procedures with three different random seeds, each achieving similar test log-likelihoods, and show the mean $\pm$ standard deviation across seeds as shaded regions. 
(A) follows the plotting convention of Fig.~\ref{fig:main_ibl2}A, showing stimulus- and choice-dependent weight updates. 
(B) follows the plotting convention of Fig.~\ref{fig:main_ibl2}B, showing reward-history–conditioned updates. 
While some variation is present, the essential qualitative features—such as side bias, negative baseline, and reward-history dependence—are consistently recovered across seeds. 
} \label{fig:supp_shaded}
\end{figure}

\begin{figure}[h!]
    \centering
    \includegraphics[width=0.9\textwidth]{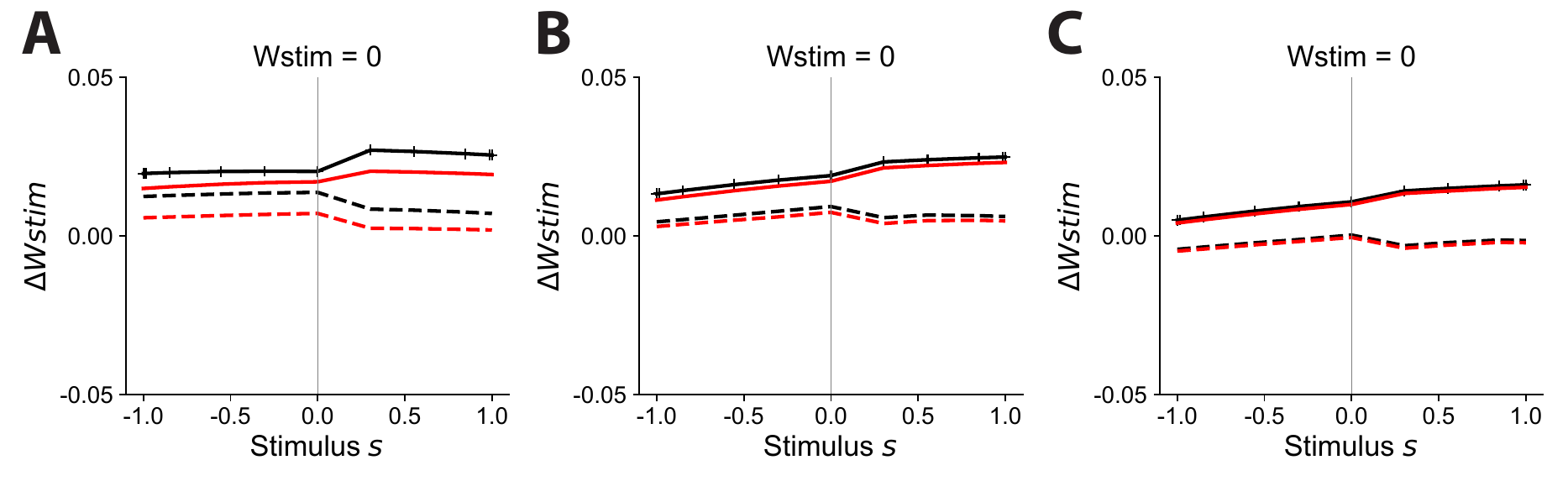}
    \caption{\textbf{Longer history effects observed in the model.} 
In Fig.~~\ref{fig:main_ibl2}B, we conditioned on up to four consecutive rewarded or unrewarded trials, since more consecutive trials risk extrapolation. 
Here we bypass this issue by still conditioning on four consecutive rewarded/unrewarded trials, but taken from further back in the trial history, to test whether the model can capture dependencies beyond the immediate past. 
(A) conditioned on trials 7 to 10 trials ago, (B) 17 to 20 trials ago, and (C) 27–30 trials ago. 
History effects are still detectable beyond just four trials, but decay with distance. Plotting conventions are the same as in Fig.~\ref{fig:main_ibl2}B.}
\label{fig:supp_toffset} 
\end{figure}

\newpage 

\newpage 

\begin{figure}[h!]
    \centering
    \includegraphics[width=0.9\textwidth]{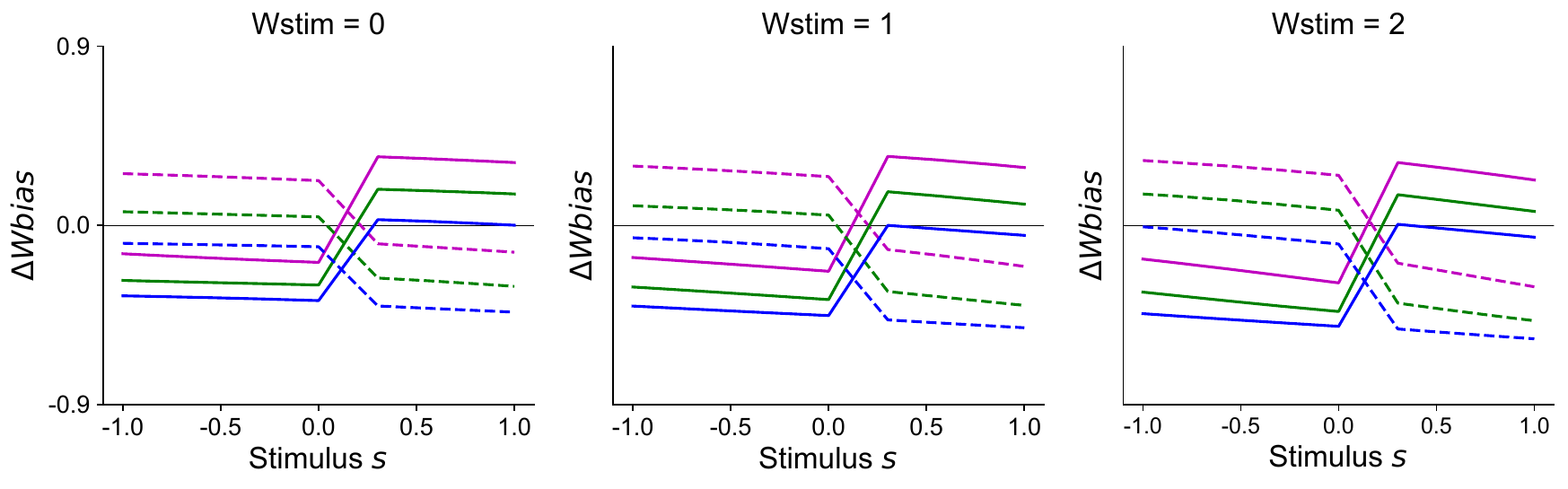}
    \caption{ The inferred update function for the bias weight, using RNNGLM, qualitatively matches the fitting in Fig.~\ref{fig:supp_IBL_db}. }
 \label{fig:supp_IBL_db_rnn}
\end{figure} 

\begin{figure}[h!]
    \centering
    \includegraphics[width=0.98\textwidth]{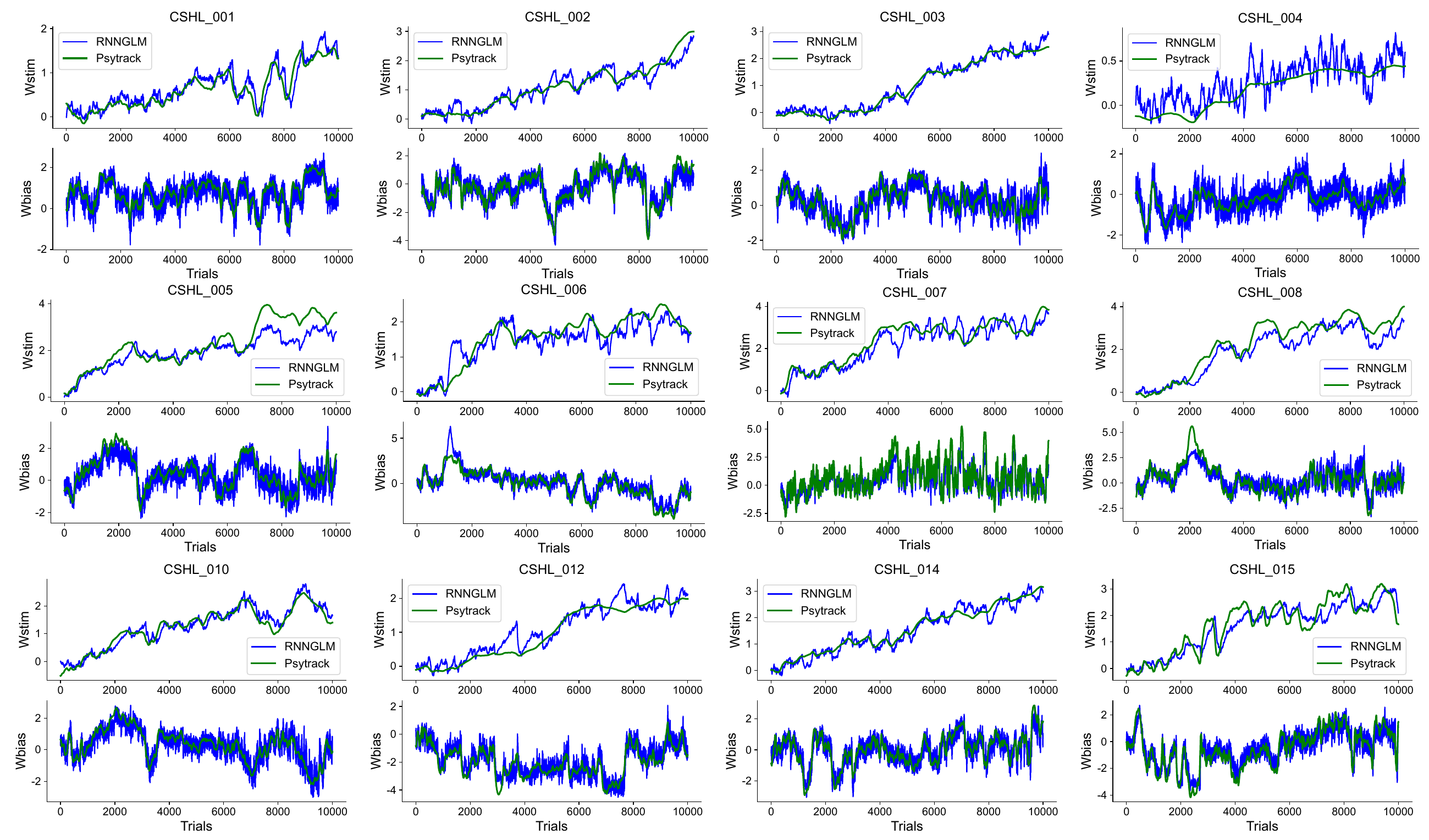}
    \caption{ Similar to Fig.~\ref{fig:supp_psyDNN} but for RNNGLM, where recovered weights also closely resemble those inferred by PsyTrack. 
} \label{fig:supp_psyRNN}
\end{figure} 

\end{document}